\documentclass[10pt]{article} %
\usepackage[accepted]{rlc}

\usepackage{amssymb}            %
\usepackage{mathtools}          %
\usepackage{mathrsfs}           %
\usepackage{graphicx}           %
\usepackage{subcaption}         %
\usepackage[space]{grffile}     %
\usepackage{url}                %
\usepackage{multicol}

\usepackage{color}

\title{Online Planning in POMDPs with State-Requests}

\author{%
  Rapha\"el Avalos$^{1, 2}$\thanks{Work done during a research visit at TU Delft.}
  \quad Eugenio Bargiacchi$^{1}$ \quad
  {Ann Now\'e}$^{1\ }$\\
  \textbf{Diederik M.\ Roijers$^{1,3}$} \quad
  \textbf{Frans A.\ Oliehoek$^{2}$}\\
  $^1$ AI Lab, Vrije Universiteit Brussel \quad $^2$ TU Delft \quad
  $^3$ City of Amsterdam  \\
  \texttt{raphael.avalos@vub.be}
}

\usepackage{amsmath,amsfonts,bm, amsthm, amssymb}
\usepackage{mathabx}
\usepackage{dsfont}
\usepackage{nicefrac}

\makeatletter
\newcommand{\pushright}[1]{\ifmeasuring@#1\else\omit\hfill$\displaystyle#1$\fi\ignorespaces}
\newcommand{\pushleft}[1]{\ifmeasuring@#1\else\omit$\displaystyle#1$\hfill\fi\ignorespaces}
\makeatother

\newtheorem{theorem}{Theorem}%
\newtheorem{definition}[theorem]{Definition}

\newtheorem{lemma}[theorem]{Lemma}

\theoremstyle{remark}
\newtheorem{remark}{Remark}

\newenvironment{proofsketch}{%
  \proof}{\endproof}

\newcommand{\tuple}[1]{\ensuremath{\left\langle #1 \right\rangle}}

\newcommand{\fun}[1]{\ensuremath{\mathopen{}\mathclose\bgroup\left(#1\aftergroup\egroup\right)}}

\newcommand{\mdp}{\ensuremath{\mathcal{M}}}
\newcommand{\states}{\ensuremath{\mathcal{S}}}
\newcommand{\actions}{\ensuremath{\mathcal{A}}}

\newcommand{\probtransitions}{\ensuremath{\mathbf{P}}} %
\newcommand{\rewards}{\ensuremath{\mathcal{R}}}

\newcommand{\state}{\ensuremath{s}}

\newcommand{\action}{\ensuremath{a}}

\newcommand{\reward}{\ensuremath{r}}

\newcommand{\act}[1]{\ensuremath{\mathit{Act}\ifthenelse{\equal{#1}{}}{}{(#1)}}}

\newcommand{\policy}{\ensuremath{\pi}}

\newcommand{\stationary}[1]{\ensuremath{\xi_{#1}}}

\newcommand{\pomdp}{\ensuremath{\mathcal{P}}}
\newcommand{\observations}{\ensuremath{\Omega}}
\newcommand{\observationfn}{\ensuremath{\mathcal{O}}}
\newcommand{\observation}{\ensuremath{o}}
\newcommand{\pomdptuple}{\langle \states, \observations, \actions, \probtransitions, \observationfn, \rewards,  \discount \rangle}
\newcommand{\actionfn}{\ensuremath{\mathbf{A}}}

\newcommand{\histories}{\ensuremath{\fun{\actions \cdot \observations}^{*}}}

\newcommand{\history}{\ensuremath{h}}

\newcommand{\belief}{\ensuremath{b}}
\newcommand{\beliefs}{\ensuremath{\mathcal{B}}}

 \newcommand{\encoderparameter}{\ensuremath{}}

\newcommand{\discount}{\ensuremath{\gamma}}

\newcommand{\expectedsymbol}[1]{\ensuremath{\mathop{\mathbb{E}}\ifthenelse{\equal{#1}{}}{}{_{#1}}}}

\newcommand{\normal}[3]{\ensuremath{\displaystyle \ifthenelse{\equal{#3}{}}{\mathcal{N}(#1, #2)}{\mathcal{N}(#3\,;\, #1, #2)}}}

\newcommand{\overbar}[1]{\mkern 1.5mu\overline{\mkern-1.5mu#1\mkern-1.5mu}\mkern 1.5mu}
\newcommand{\overbarit}[1]{\,\overline{\!{#1}}}
\newcommand{\embed}{\ensuremath{\phi}}

\newcommand{\latentprobtransitions}{\ensuremath{\overbar{\probtransitions}}}

\newcommand{\latentrewards}{\ensuremath{\overbarit{\rewards}}}

\newcommand{\latentbeliefupdate}{\ensuremath{\overbar{\tau}}}

\newcommand{\localtransitionloss}[1]{L_{\probtransitions}}
\newcommand{\localrewardloss}[1]{L_{\rewards}}
\newcommand{\observationloss}[1]{\ensuremath{L_{\observationfn}}}
\newcommand{\beliefloss}[1]{\ensuremath{L_{\latentbeliefupdate}}}
\newcommand{\onpolicyrewardloss}[1]{\ensuremath{L_{\latentrewards}^{\varphi}}}
\newcommand{\onpolicytransitionloss}[1]{\ensuremath{L_{\latentprobtransitions}^{\varphi}}}

\newcommand{\KR}[1]{\ensuremath{\ifthenelse{\equal{#1}{}}{K_{\latentrewards}}{K_{\latentrewards}^{#1}}}}
\newcommand{\KP}[1]{\ensuremath{\ifthenelse{\equal{#1}{}}{K_{\latentprobtransitions}}{K_{\latentprobtransitions}^{#1}}}}

\newcommand{\originaltolatentstationary}[1]{{\latentprobtransitions_{\embed_{\encoderparameter}\stationary{\ifthenelse{\equal{#1}{}}{\policy}{#1}}}}}

\def\1{\bm{1}}

\DeclareMathAlphabet{\mathsfit}{\encodingdefault}{\sfdefault}{m}{sl}
\SetMathAlphabet{\mathsfit}{bold}{\encodingdefault}{\sfdefault}{bx}{n}

\newcommand{\pomdpsr}{\ensuremath{\pomdp_{\text{SR}}}}
\newcommand{\cost}{\ensuremath{c}}
\newcommand{\pomdpsrtuple}{\ensuremath{\langle \pomdp, \cost \rangle}}
\newcommand{\request}{\ensuremath{\iota}}
\newcommand{\notrequest}{\ensuremath{\bar{\iota}}}
\newcommand{\cumProb}{\ensuremath{\Psi}}
\newcommand{\cumProbDirect}{\ensuremath{\bar{\Psi}}}
\newcommand{\pathSet}{\ensuremath{\Phi}}
\newcommand{\fringe}{\ensuremath{\mathcal{F}}}
\newcommand{\graph}{\ensuremath{\mathcal{G}}}
\newcommand{\tree}{\ensuremath{\mathcal{T}}}
\newcommand{\support}{\ensuremath{\text{supp}}}
\newcommand{\origin}{\ensuremath{\xi}}
\newcommand{\rootBelief}{\ensuremath{\belief_{0}}}
\newcommand{\gap}[1]{\ensuremath{e\fun{#1}}}
\newcommand{\gapApprox}[2]{\ensuremath{\hat{e}_{#1}\fun{#2}}}

\DeclareMathOperator*{\argmax}{arg\,max}

\newcommand{\smallparagraph}[1]{\smallskip\noindent\textbf{#1}}
\usepackage[inline]{enumitem}

\usepackage{multirow}
\usepackage{booktabs}
\usepackage{subcaption}

\usepackage{graphicx} 
\usepackage{tabularx}
\usepackage{subcaption}
\usepackage[ruled,vlined,linesnumbered]{algorithm2e}

\usepackage{caption}
\usepackage{subcaption}
\usepackage{todonotes}
\usepackage{wrapfig}
\usepackage{bbm}

\begin{document}

\maketitle

\begin{abstract}
In key real-world problems, full state information is sometimes available but only at a high cost, like activating precise yet energy-intensive sensors or consulting humans, thereby compelling the agent to operate under partial observability. For this scenario, we propose AEMS-SR (Anytime Error Minimization Search with State Requests), a principled online planning algorithm tailored for POMDPs with state requests. By representing the search space as a graph instead of a tree, AEMS-SR avoids the exponential growth of the search space originating from state requests.  
Theoretical analysis demonstrates AEMS-SR's $\varepsilon$-optimality, ensuring solution quality, while empirical evaluations illustrate its effectiveness compared with AEMS and POMCP, two SOTA online planning algorithms. 
AEMS-SR enables efficient planning in domains characterized by partial observability and costly state requests offering practical benefits across various applications.
\end{abstract}

\section{Introduction}

The \textit{Partially Observable Markov Decision Process (POMDP)} is a powerful framework that models sequential decision-making in scenarios where the environment's true state is inaccessible.
Often, this partial observability is an inherent characteristic of the environment, perhaps due to noise or the unavailability of suitable sensors. 
However, in numerous instances, determining the true state of the system is feasible but entails a considerable cost. 
For example, consider a scenario involving a battery-powered robot that lacks the necessary power to employ highly accurate sensors continuously, and thus, is also equipped with power-efficient yet less precise sensors. 
Additionally, the concept of requesting state information can extend to scenarios with privacy implications, such as the use of surveillance cameras in public spaces for crowd control. In these cases, the decision to activate cameras involves weighing the benefits of state access against potential privacy costs.

We can think of these settings as situations where the agent has the option to consult an oracle (like a precise sensor or a human expert) at every step to obtain the state against a cost.
In the context of our battery-powered robot, the cost could represent the electricity cost of activating the accurate sensor. 
We refer to this setting as \textit{POMDPs with State Requests (POMDP-SR)}, where the agent, for a cost, can eliminate all uncertainty regarding its current state before selecting each action.

A naive approach to handling POMDP-SRs is converting them to equivalent POMDPs, as detailed in Section \ref{sec:framework}.
However, such a method overlooks the unique characteristics of POMDP-SRs, potentially leading to suboptimal performance of conventional POMDP planning techniques. This is largely because in the conversion to an equivalent POMDP, the number of time-steps effectively doubles, and the observation space expands considerably.
For methods relying on tree search, such as POMCP \citep{POMCP} and AEMS \citep{AEMS}, 
the transformation into an equivalent POMDP introduces an exponential increase in the search tree, significantly impeding their efficiency.
Extensions that build on sparse samplings, such as DESPOT \citep{somani2013despot}, theoretically can visit only a small part of the tree, but to obtain good results, this small part is in practice still large.

In this paper, we introduce a novel online planning algorithm, AEMS-SR (Anytime Error Minimization Search with State Requests), tailored for POMDPs with state requests. 
While traditional approaches use trees, AEMS-SR can leverage a cyclic graph, significantly reducing the search space by avoiding redundant expansions and improving computational efficiency.

Our contributions are threefold: 1) We formalize the request the state framework; 2) We introduce a new algorithm, AEMS-SR, and theoretically demonstrate its $\varepsilon$-optimality; 3) We conduct experiments on RobotDelivery, our newly developed benchmark, and Tag demonstrating AEMS-SR's superiority over AEMS and POMCP. Our results highlight AEMS-SR's efficiency in circumventing the exponential growth of the search tree, highlighting its potential in this challenging setting.

\section{Background}

\smallparagraph{POMDPs}~ 
Partially Observable Markov Decision Processes \citep{aastrom1965optimal} are defined as a tuple $\pomdp = \pomdptuple$ where $\states$ is the set of states, $\observations$ is the sets of observations, $\actions$ is the set of actions, $\probtransitions \colon \states \times \actions \rightarrow \Delta_\states$ is probability transition function with $\Delta_\states$ being the simplex over the state space,  $\observationfn \colon \states \times \actions \rightarrow \Delta_\observations$ is the probability observation function, $\rewards \colon \states \times \actions \rightarrow \mathbb{R}$ is the reward function, and $\gamma \in [0, 1)$ is the discount factor. At each time step the agent selects an action based on its observation-action history $\history \in \histories$.
Due to the exponential growth of the history space in the number of time-steps dealing with histories might not be practical.
Beliefs, defined as the probability distribution over current states $\belief \in \beliefs \equiv \Delta_\states$, are sufficient statistics of the history for control \citep{aastrom1965optimal} and a more compact alternative. Beliefs are computed recursively: after taking an action $a$ in belief $b$ and receiving an observation $o$ the next belief is defined for any next state $\state' \in \states$ as follows, with $\eta$ being the normalizing factor.
\begin{align}
    \belief'(\state') = \eta \expectedsymbol{\state \sim \belief} \probtransitions\fun{\state' \mid \state, \action} \cdot \observationfn\fun{\observation \mid \state', \action}
    \notag
\end{align}
A policy $\policy \colon \beliefs \rightarrow \actions$ is a mapping from beliefs to actions and is associated with a Value $V^\policy\fun\belief$.  We denote as $\pi^*$ and $V^*$ the optimal policy and its value.
For finite horizon, $V$ is a piece-wise linear and convex (PWLC) function of the belief \citep{sondik1971optimal}, and can therefore be represented as a set $\Gamma$ of $\alpha$-vectors which corresponds to the slopes of the PWLC function. 

We refer to beliefs as \textit{corner beliefs} when the probability of being in a state $\state$ is 1 and 0 for the other states.  In clear contexts, we directly use the state $\state$ to reference such beliefs.
The support of a belief, $\support\fun{\belief}$, is the set of states with non-zero probability.
POMDP planning methods generally fall into two categories: offline and online approaches. 
Offline methods precompute comprehensive plans for all scenarios but suffer from computational demands and scalability issues. In contrast, online methods provide real-time computational capabilities for determining optimal actions within time constraints.

\smallparagraph{AEMS} 
Anytime Error Minimization Search (AEMS) \citep{AEMS} is an online algorithm that, following the stochastic shortest path approach of AO* \citep{nilsson1982principles}, builds a tree $\tree$ from the current belief $\rootBelief$.
The algorithm maintains an upper bound $U_{\tree}(\belief)$ and a lower bound $L_{\tree}(\belief)$ of the value $V^*\fun{\belief}$.
At each step, AEMS expands the node that is believed to have the highest reduction potential for the error at the root.
Let $\fringe(\tree)$ be the set of fringe nodes (nodes without children) in $\tree$, $\hat{e}(\belief) = U(\belief) - L(\belief)$ be the gap between the upper and lower bounds of the value, $d_{\tree}(\belief, \rootBelief)$ be the number of actions that separate $\belief$ and $\rootBelief$ in the tree $\tree$, $\history_{\rootBelief}^\belief$ be the history from $\rootBelief$ to $\belief$, and $P(\history_{\rootBelief}^\belief \mid \rootBelief, \hat{\policy}_\tree)$ be the probability of reaching $\belief$ from $\rootBelief$ by following the policy $\hat{\policy}_\tree$ that selects the action maximizing the upper bound. AEMS expands the fringe node that maximizes the heuristic of Eq.~\ref{eq:aems}. 
\begin{align}
    \Tilde{\belief}(\tree) &= \argmax_{\belief \in \fringe(\tree)} \gamma^{d_{\tree}(\belief, \rootBelief)} P(\history_{\rootBelief}^\belief \mid \rootBelief, \hat{\policy}_\tree) \hat{e}(\belief)
\label{eq:aems}
\end{align}

POMCP \citep{POMCP} extends MCTS to POMDPs. The algorithm relies on rollouts and removes the need to compute belief updates allowing it to scale to large state spaces.

\section{Framework}
\label{sec:framework}

\begin{figure*}[t]
    \centering
    \begin{subfigure}{.6\textwidth}
      \centering
      \includegraphics[width=\linewidth]{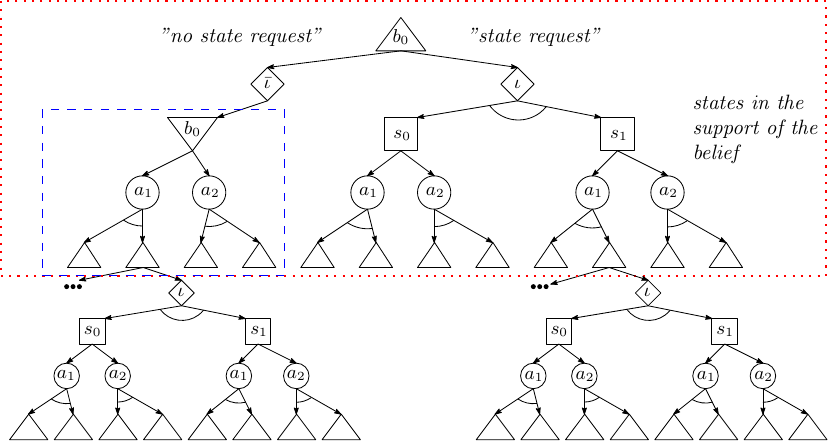}
      \caption{AO Tree. The red box corresponds to an expansion in a $\pomdpsr$. For comparison, the blue box would match the expansion in a $\pomdp$. }
      \label{fig:sub1}
    \end{subfigure}%
    \begin{subfigure}{.4\textwidth}
      \centering
      \includegraphics[width=.9\linewidth]{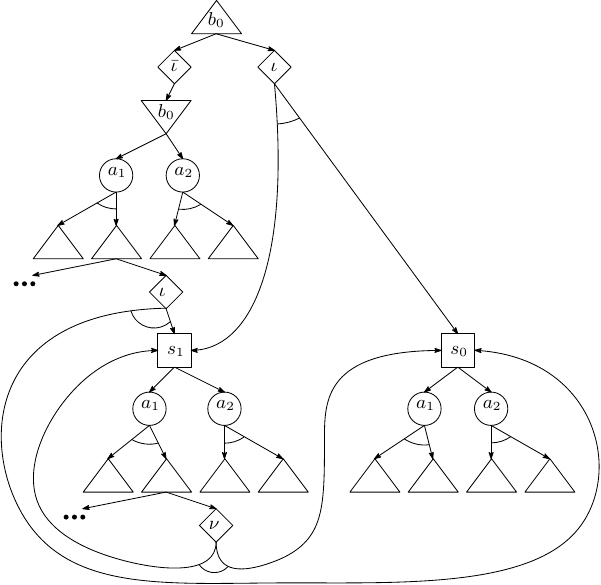}
      \caption{AO Graph}
      \label{fig:sub2}
    \end{subfigure}%
    \caption{Tree and Graph representation after three successive expansions (the expanded beliefs are in green). 
    Beliefs before selecting state request depicted by upward triangles, beliefs before environmental action by downward triangles, (not-)request state actions by diamonds, environmental actions by circles, and corner beliefs by rectangles. 
    Some nodes are hidden for readability.
    }
\end{figure*}

\smallparagraph{POMDP-SR}~ 
We define the POMDP with State Request as a tuple $\pomdpsr = \pomdpsrtuple$ where $\pomdp$ is a POMDP and $\cost > 0$ is the associated cost to request the state.
At each timestep, the agent first decides whether to request the state, which is immediately revealed if requested and then selects an action. The decision is binary: $\request$ to request and $\notrequest$ to not request the state.

A property of  POMDP-SR that may not be immediately apparent is that even in cases where the optimal actions in an MDP and a POMDP align, the POMDP-SR's optimal action might differ. This arises from the fact that the agent operates under the anticipation of potential future state requests.  
In such contexts, a suboptimal action in the MDP and the POMDP can become the optimal one in the POMDP-SR when combined with a future request the state, achieving a return that is sub-optimal for the MDP but significantly better than the one of a POMDP.   
An illustrative example demonstrating this aspect of POMDP-SR is elaborated in the Appendix. This highlights how the integration of state requests fundamentally shifts the dynamics of decision-making in POMDPs.

\smallparagraph{Equivalent POMDP}~
A POMDP-SR $\pomdpsr = \pomdpsrtuple$ can be transformed into an equivalent POMDP $\pomdp' = \langle \states', \observations', \actions', \actionfn', \probtransitions', \observationfn', \rewards',  \discount' \rangle$  with variable action space and with $\probtransitions', \observationfn', \rewards'$ only defined over legal actions.  While the comprehensive technicalities of this transformation are detailed in the Appendix, the core concept is to separate the state request action from the environmental action doubling the number of timesteps. 
Additionally, the state space is expanded by integrating a binary indicator, which functions to signal the phase in which the agent is operating.
\footnote{
Including the state request action $\request$ as an additional action at every step avoids doubling timesteps and state space, but does not yield an equivalent model to a POMDP-SR due to the discounting factor.}
For any state $\state \in \states$, $\state^0$ indicates the request the state phase, and $\state^1$ is the environment action phase. We denote similarly the beliefs containing only states of one type (i.e. $\belief^0, \belief^1).$

\smallparagraph{Equivalent POMDP Complexity}~
Transforming a POMDP-SR into its equivalent POMDP enables the use of classic POMDP planning algorithms, but this approach may prove inefficient. 
One inefficiency arises from the lack of support for variable action spaces in classic implementations, necessitating the use of deterrent penalties for illegal action  
which impacts the algorithm's complexity. 
For offline methods like PBVI, doubling the state space and augmenting the observation space to include the state space lead to a steep increase in complexity. Heuristic search algorithms like AEMS face an even more challenging situation as doubling the horizon in POMDP-SR and requiring a new sub-tree for every state in the support of the belief result in an exponential expansion of the search tree (Fig.~\ref{fig:sub1}). This growth underscores the fundamental issue: current planning algorithms are ill-suited to effectively handle the unique complexities introduced by POMDP-SR scenarios.

\section{Online planning: AEMS-SR}

In this section, we present our new method Anytime Error Minimization Search for POMDP-SRs (AEMS-SR) which adapts AEMS to our framework. 
As outlined in section \ref{sec:framework}, the introduction of state requests leads to an exponential increase in the size of the search tree. This growth is primarily due to two factors: the doubling of timesteps and the generation of a new subtree for each state in the belief's support. Consequently, each expansion of belief in a POMDP-SR, illustrated by the red box in Fig.~\ref{fig:sub1}, adds $(1 + |\support(\belief)|) \cdot |\actions| \cdot |\observations|$ nodes. %
This is in stark contrast to classic POMDPs, where only $|\actions| \cdot |\observations|$ nodes are added per expansion, as illustrated by the blue box in Fig.~\ref{fig:sub1}.

Upon examining the search tree in POMDP-SR scenarios, illustrated in Figure \ref{fig:sub1}, we observe that many nodes are similar. This redundancy is particularly pronounced in cases involving position uncertainty and potential action failure, leading to a significant overlap in subsequent beliefs. As a result, the tree often contains identical subtrees that are redundantly expanded, impairing search efficiency.
The challenge of repetitive subtree expansions is not unique to POMDP-SR; it is a known issue in both POMDPs and MDPs. Techniques like transposition tables \citep{childs2008transpositions} have been used to address this problem, offering computational trade-offs that can be beneficial in certain environments but are less practical for continuous spaces such as beliefs.
To deal with these, AEMS-SR employs a rooted cyclic graph, denoted as $\graph$, with the current belief $\rootBelief$ as its root, for the search replacing the conventional tree structure. {A rooted cyclic graph is defined as a regular cyclic graph where every node can be reached from its root, and where the root does not have any parents.} This shift to a cyclic graph necessitates the development of novel heuristic and algorithmic solutions to adeptly manage the added complexities of cyclicity.

\subsection{AEMS-Loop}

We first introduce AEMS-Loop, the extension of AEMS to cyclic graphs, and theoretically prove its completeness and $\varepsilon-$optimiality, meaning that the algorithm will always return a solution that is $\varepsilon-$close to the optimal solution given enough time. 
Similar to other online tree search algorithms, we rely on upper and lower bounds, denoted as $U\fun{\belief}$ and $L\fun{\belief}$, of the optimal value function $V^*\fun{\belief}$ that are computed offline. These values are propagated in the graph $\graph$ to the parents using the following equations, allowing expansions to reduce the error gap at the root: $e_\graph\fun{\rootBelief} = U_\graph\fun{\rootBelief} - L_\graph\fun{\rootBelief}$.
\begin{align}
     U_\graph\fun{\belief, \action} &= R\fun{\belief, \action} + \gamma \sum_{\observation \in \observations} P\fun{\observation \mid \belief, \action} U_\graph\fun{\tau\fun{\belief, \action, \observation}}& \quad& U_\graph\fun{\belief} = 
        \begin{cases}
          U\fun{\belief} & \text{if $\belief \in \fringe\fun\graph$}\\
          \max_{\action \in \actions} U_\graph\fun{\belief, \action} & \text{otherwise}
        \end{cases}
        \label{eq:upper_belief}\\
    L_\graph\fun{\belief, \action} &= R\fun{\belief, \action} + \gamma \sum_{\observation \in \observations} P\fun{\observation \mid \belief, \action} L_\graph\fun{\tau\fun{\belief, \action, \observation}}& \quad& L_\graph\fun{\belief} = 
        \begin{cases}
          L\fun{\belief} & \text{if $\belief \in \fringe\fun\graph$}\\
          \max_{\action \in \actions} L_\graph\fun{\belief, \action} & \text{otherwise}
        \end{cases}
        \label{eq:lower_belief}
\end{align}

Working with a cyclic graph introduces the possibility of multiple paths, and potentially an infinite number, between the root $\rootBelief$ and any fringe node $\belief \in \fringe\fun\graph$. 
We define as $\pathSet_\graph\fun{\rootBelief, \belief}$ the set of paths in $\graph$ that start on $\rootBelief$ and end up on $\belief$. A path $\history \in \pathSet_\graph\fun{\rootBelief, \belief}$ is a sequence of beliefs, action and observation $(\belief_i, \action_i, \observation_{i+1})_{i < T}$. 
Based on a policy $\pi$, each path has an associated probability $P\fun{\history | \rootBelief, \pi} = \Pi_{i=0}^{T - 1} P\fun{\observation_{i+1} | \belief_i, \action_i} \pi\fun{\action_i | \belief_i}$ corresponding to the probability of observing the path $\history$ while starting from $\rootBelief$ and following $\pi$. 
Additionally, we define $\cumProb^\graph_{\pi}\fun{\rootBelief, \belief}$ as the sum over all possible paths between the root $\rootBelief$ and a fringe $\belief \in \fringe\fun{\graph}$ of the probability of observing the path discounted by the length of the path $d(\history)$. We drop the subscript $\graph$ when the dependency is clear.
\begin{align}
    \cumProb^\graph_{\pi}\fun{\rootBelief, \belief}= \sum_{\history \in \pathSet_\graph\fun{\rootBelief, \belief}} \gamma^{d\fun{\history}} \probtransitions\fun{\history | \rootBelief, \policy }
    \label{eq:cum_prod_def}
\end{align}

\begin{theorem}
    In any rooted graph $\graph$ with root $\rootBelief$ where values are computed according to Eq. \ref{eq:lower_belief} using a lower bound value function L with error $e(\belief) = V^*(\belief) - L(b)$, the error on the root belief state is bounded by: 
    $e_\graph(\rootBelief) = V^*(\rootBelief) - L_\graph(\rootBelief) \leq \sum_{\belief \in \fringe\fun\graph} \cumProb_{\pi^*}\fun{\rootBelief, \belief} e(\belief)$ where $e(\belief) = V^*(\belief) - L(\belief)$.
    \label{th:th1}
\end{theorem}
\begin{proofsketch}
    We use a similar proof as AEMS on an enrolling of the tree of size $n$ to obtain an upper bound composed of two elements: (i) the discounted probabilities of observing a path of a size at most $n$ from the root to one of the replicas of an element in $\fringe\fun\graph$; (ii) the discounted probabilities of other paths which are of size $n$. By making $n\rightarrow+\infty$, the first part of the upper-bound converges to the term in the theorem and the second to 0.
\end{proofsketch}
Theorem \ref{th:th1} gives an upper bound on the contribution of each fringe node to the error at the root, extending AEMS's Theorem 1 \citep{AEMS} to cyclic graphs. 
Assuming a tree structure, which implies $|\pathSet\fun{\rootBelief, \belief}| = 1$ for any fringe belief $\belief \in \fringe\fun{\tree}$, recovers the original theorem.

Similar to AEMS, this theorem provides a robust method for choosing the next belief to expand to rapidly minimize root error: prioritize expanding the belief with the greatest estimated contribution $\arg\max_{\belief \in \fringe\fun\graph} \cumProb_{\pi^*}\fun{\rootBelief, \belief} e(\belief)$.
However, as the optimal policy $\pi^*$ and value function $V^*$ are unknown, we need to approximate them to compute $\cumProb_{\pi^*}\fun{\rootBelief, \belief}$  and $e(\belief)$. We denote the approximation of $\pi^*$ as $\hat{\pi}_\graph$. As in AEMS, we employ the following two approximations:\vspace{-\baselineskip}\\
\begin{minipage}[t]{0.5\linewidth}
\begin{align}
    \label{eq:policy_t}
    \hat{\pi}_\graph\fun{\belief, \action} = \mathbbm{1}\{a = \arg\max_{\action'} U_\graph\fun{\belief, \action}\} 
\end{align}
\end{minipage}
\hfill
\begin{minipage}[t]{0.5\linewidth}
\begin{align}
    \label{eq:error_approximation}
    \gapApprox{}{\belief} = U\fun{\belief} - L\fun{\belief} \geq e\fun{\belief}
\end{align}
\end{minipage}

While other approximations $\hat{\pi}_\graph$ are possible, we selected the one presented in Equation \ref{eq:policy_t} because of its empirical performance in AEMS and its simplicity. %
Using those two approximations, we can leverage Theorem 1 to define the following heuristic for selecting the next belief to expand $\Tilde{\belief}\fun{\graph}$:
\begin{align}
    \label{eq:heurestic}
    \Tilde{\belief}\fun{\graph} = \arg\max_{\belief \in \fringe\fun{\graph}} \cumProb_{\hat{\pi}_\graph}\fun{\rootBelief, \belief} \gapApprox{}{\belief}
\end{align}
\begin{theorem}
    Given $U$ bounded above, $L$ bounded below such as $\forall \belief \in \beliefs$, $U\fun{\belief} \geq V^*\fun{\belief} \geq L\fun{\belief}$, and $\gapApprox{}{\belief} = U\fun{\belief} - L\fun{\belief}$ , if $\gamma \in [0, 1)$ and $\inf_{\belief, \graph | \gapApprox{\graph}{\belief} > \varepsilon} \hat{\pi}_\graph\fun{\belief, \hat{\action}_\belief^\graph} > 0$ for $\hat{\action}_\belief^\graph = \arg\max_{\action \in \actions} U_\graph\fun{\belief, \action}$, then the AEMS-Loop algorithm using heuristic $\Tilde{\belief}\fun{\graph}$ is complete and $\varepsilon-$optimal.
    \label{th:th2}
\end{theorem}

Theorem \ref{th:th2} establishes the completeness and $\varepsilon-$optimality of AEMS-Loop for any policy $\hat{\pi}_\graph$ assigning non-zero probability to the upper-bound maximizing action, such as the one defined in Eq.\ref{eq:policy_t}.

\subsection{Algorithm}

This subsection explains how AEMS-SR (Alg.~\ref{algo:aems-sr}), a practical implementation of AEMS-Loop adapted to POMDP-SR, works in practice.
While the graph structure enables consolidating identical belief nodes to prevent redundant work, fully implementing this strategy would require comparing each new belief against all existing beliefs in the current graph. Such an approach would lead to scalability issues similar to those that render graphs less practical in general MDPs and POMDPs. 
To maintain tractability while still achieving our main goal of mitigating the exponential growth in the search tree, we restrict the capacity for multiple parents to corner beliefs. 

AEMS-SR starts with a graph $\graph$ containing only the root belief $\rootBelief$. The algorithm then iterates through the following steps until the time limit is reached: a) find the fringe belief $\Tilde{\belief}\fun{\graph} \in \fringe\fun\graph$ that maximizes Eq.~\ref{eq:heurestic}, this corresponds to  Alg.~\ref{alg:getbelieftoexpand}; b) update the graph $\graph$ by expanding the belief $\Tilde{\belief}\fun{\graph}$, for the request the state action $\request$ we reuse the existing corner beliefs creating a graph similar to Fig.~\ref{fig:sub2}; c) update the upper and lower bound value to match Eq.\ref{eq:upper_belief}-\ref{eq:lower_belief}.

For all fringe beliefs $\belief \in \fringe\fun\graph$, we define the \textit{origin} function $\origin_\graph\fun{\belief}$, which returns the first ancestor among the corner beliefs, or $\rootBelief$ if such an ancestor does not exist. Additionally, we define $\history^{\origin}_\belief$ as the unique path between the origin $\origin\fun{\belief}$ and the belief $\belief$ which does not contain a state request action $\request$. The fact that $\belief \in \fringe\fun\graph$ ensures the existence of the path, while the uniqueness is guaranteed by construction as only the corner beliefs can have multiple parents.

\smallparagraph{Computing $\cumProb$}~
Our heuristic (Eq. \ref{eq:heurestic}) requires to compute $\cumProb\fun{\rootBelief,\belief}$ (Eq. \ref{eq:cum_prod_def}) for all fringe beliefs $\belief \in \fringe\fun\graph$.
While straightforward in a tree structure, it becomes complex in a graph due to potentially infinite paths 
between $\rootBelief$ and $\belief$. 
We address this challenge by leveraging the fact that only corner beliefs can have multiple parents 
to obtain Eq. \ref{eq:cumprod}. As the second part of the equation is straightforward to compute, our focus shifts to calculating $\cumProb\fun{\rootBelief, \state}$ for all corner beliefs $\state$.
\begin{align}
    \cumProb\fun{\rootBelief, \belief} = \cumProb\fun{\rootBelief, \origin\fun\belief} P\fun{\history^\origin_\belief | \origin\fun{\belief}, \hat{\pi}_\graph}
    \label{eq:cumprod}
\end{align}

We start by reducing the graph $\graph$, such as the one in Figure \ref{fig:sub2}, 
to $\bar{\graph}$ containing only the root belief and corner beliefs.
The set of nodes of $\bar{\graph}$ is defined as $N'=\{\rootBelief\}\cup\states$. An edge connects a node $\belief \in N'$ to a corner belief $\state \in \states$ if there is at least one \textit{direct path} $\history$ between the two nodes in the original graph $\graph$, with a direct path defined as a path where only the last action is a request the state $\request$. 
This edge is weighted by the sum of the discounted cumulative probabilities over direct paths:
\begin{align*}
    \text{edge}\fun{\belief, \state} = 
    \sum_{\substack{\history \in \pathSet\fun{\belief, \state}\\  \history \text{ is direct}}} \gamma^{d\fun{\history}} \probtransitions\fun{\history | \belief, \policy_\graph }
\end{align*}
For nodes $n, n' \in N'$, we define $\cumProbDirect\fun{n, n'}$ as 0 if there is no edge between $n$ and $n'$, and as the weight of the edge otherwise.
As the paths starting in $\rootBelief$ and ending in a corner belief $\state$ are either direct or pass through another corner belief $\state'$, we can write for all corner beliefs $\state$: 
\begin{align}
    \cumProb\fun{\rootBelief, \state} = \sum_{\state' \in \states} \cumProb\fun{\rootBelief, \state'} \cumProbDirect\fun{\state', \state} + \cumProbDirect\fun{\rootBelief, \state}%
    \label{eq:cumprob_direct}
\end{align}
We can rewrite Eq. \ref{eq:cumprob_direct} more compactly into Eq.\ref{eq:cumprob_direct_compact} using matrix notations by defining $\cumProbDirect_{\state, \state'} = \cumProbDirect\fun{\state, \state'}$, $\cumProbDirect_{\rootBelief}$ as the vector with components equal to $\cumProbDirect\fun{\rootBelief, \state}$, and $\cumProb_{\rootBelief}$ the vector with components equal to $\cumProb\fun{\rootBelief, \state}$. From which we obtain the solution given in Eq. \ref{eq:cum_prod_inverse}.\vspace{-\baselineskip}\\
\begin{minipage}[t]{0.5\linewidth}
\begin{align}
    \cumProb_{\rootBelief} &= \cumProbDirect \cumProb_{\rootBelief} + \cumProbDirect_{\rootBelief} 
    \label{eq:cumprob_direct_compact}
\end{align}
\end{minipage}
\hfill
\begin{minipage}[t]{0.5\linewidth}
\begin{align}
    \cumProb_{\rootBelief} &= (I - \cumProbDirect)^{-1} \cumProbDirect_{\rootBelief}
    \label{eq:cum_prod_inverse}
\end{align}
\end{minipage}

We note that this solution could be seen as constructing the Markov chain corresponding to the corner beliefs and finding the stationary distribution of the policy $\hat{\pi}_\graph$.
We can now compute $\cumProb\fun{\rootBelief, \belief}$ for all fringes $\belief \in \fringe\fun\graph$ (Eq. \ref{eq:cumprod}), and expand $\Tilde{\belief}\fun\graph$ (Eq. \ref{eq:heurestic}).  

Algorithm \ref{alg:getbelieftoexpand} includes the pseudo-code, where line \ref{algo:getbelieftoexpand_gwalk} calls Algorithm \ref{alg:gwalk} (see Appendix) to obtain $\cumProbDirect$ and $\cumProbDirect_{\rootBelief}$. This computation enables the calculation of $\cumProb_{\rootBelief}$ (line \ref{algo:getbelieftoexpand_inverse}) and returns $\Tilde{\belief}\fun\graph$ for expansion.

\smallparagraph{Update ancestors}
consists of leveraging the knowledge gained through a belief expansion to update the lower and upper bound in the graph $L_\graph$ and $U_\graph$ by enforcing Equations \ref{eq:upper_belief} and \ref{eq:lower_belief}. As explained in LAO* \citep{LAO}, this requires running dynamic programming. When working with stochastic trees, such as AEMS, the dynamic programming results in updating the parents sequentially until reaching the root, guaranteeing convergence in a number of steps equal to the depth of the tree. In contrast, when dealing with cyclic graphs, a full dynamic programming update is needed%
, which can be computationally expensive. To address this challenge, LAO* proposes a method to limit the number of nodes to update by focusing on a subset of nodes. In our implementation, instead of traversing the graph to determine the set of nodes to consider for dynamic programming, we update the parents recursively until the updates become smaller than a threshold ($10^{-6}$ in our experiments), and we maintain a queue to deal with the cyclic structure.

The cycle of identifying the next belief to expand, expanding it, and backtracking the lower and upper bounds continues until the predefined time limit is reached or the error gap at the root, $\hat{e}_\graph\fun\rootBelief$, becomes less than $\varepsilon$. Subsequently, the agent determines whether to request the state and selects an environmental action, by maximizing the lower bound $L_\graph$. Then the agent obtains a new observation, triggering the reinitialization of the graph with the updated belief.
\begin{minipage}[t]{0.55\linewidth}
    \begin{algorithm}[H]
\SetKwInOut{Input}{input}
\DontPrintSemicolon
\caption{AEMS-SR: Anytime Error Minimization Search with State Requests}\label{algo:aems-sr}
\Input{$t$: Maximum time, $\varepsilon$: Error threshold}

\While{not \texttt{EnvironmentTerminated()}}{
    Initialize $\graph$ with initial belief state $\rootBelief^0$ as root\;
    $t_0 \leftarrow \texttt{Time()}$\;
    \While{\texttt{Time}() - $t_0 \leq t$ and not \texttt{Solved}($\rootBelief^0$, $\varepsilon$)}{
        \tcp{Calls Algorithm~\ref{alg:getbelieftoexpand}}
        $\belief^* \leftarrow \texttt{GetBeliefToExpand}(\graph)$\;\label{alg:aems-sr-get-belief-to-expand}
        \texttt{Expand}($\belief^*$)\;\label{alg:aems-sr-expand}
        \texttt{UpdateAncestors}($\belief^*$)\;\label{alg:aems-sr-backtrack}
    }
    $\hat{\action}_0 \leftarrow \arg \max_{a \in \{\notrequest, \request\}} L_\graph(\rootBelief^0, \action)$\;
    \If{$\hat{\action}_0 = \request$}{
        $s \leftarrow \texttt{GetState}()$\\
        $\hat{\action}_1 \leftarrow \arg\max_{\action \in \actions} L_\graph(\state, \action)$
    }
    \Else{
        $\hat{\action}_1 \leftarrow \arg\max_{\action \in \actions} L_\graph(\rootBelief^1, \action)$
    }
    \texttt{DoAction}($\hat{\action}_1$); 
    $\observation \leftarrow \texttt{GetObservation()}$\;
    $\rootBelief^0 \leftarrow \tau\fun{\state, \hat{\action}_1, \observation} \texttt{ if } \hat{\action}_0 = \request \texttt{ else } \tau\fun{\rootBelief^0, \notrequest, \hat{\action}_1, \observation}$\\
    \tcp{Potentially improve the bounds}
    
}
\end{algorithm}

\end{minipage}
\hfill
\begin{minipage}[t]{0.4\linewidth}
    \begin{algorithm}[H]
\DontPrintSemicolon
\caption{\\getBeliefNodeToExpand}\label{alg:getbelieftoexpand}
\SetKwInOut{Input}{input}
\Input{Graph $\graph$, root belief $\rootBelief$}
\If{$\rootBelief \in \fringe\fun{\graph}$}{
    \KwRet $\rootBelief$\;
}

\tcc{Call to Algorithm \ref{alg:gwalk} (Appendix) that traverse the graph to compute  $\cumProbDirect, \cumProbDirect_{\rootBelief}, \fringe\fun\graph$ }
$\mathcal{V}, \cumProbDirect, \cumProbDirect_{\rootBelief}, \fringe\fun\graph \leftarrow$ \texttt{GWalk}($\rootBelief, \{\}, 0_{|\states|\times|\states|}, 0_{|\states|}, \{\}$) \label{algo:getbelieftoexpand_gwalk}\;

$\cumProb_{\rootBelief} = (I_{|\states|} - \cumProbDirect)^{-1} \cumProbDirect_{\rootBelief}$ \text{(Eq. \ref{eq:cum_prod_inverse})} \label{algo:getbelieftoexpand_inverse}\;

bestE = $-\infty$\;

\For{$\belief \in \hat{\fringe}$}{
    E = $(U(\belief) - L(\belief)) \cdot \cumProb\fun{\rootBelief, \belief}$ \textit{(Eq. \ref{eq:cumprod})}\;

    \If{E $>$ bestE}{
        bestE = E\;
        $\belief_{\text{best}} = \belief$\;
    }
}

\KwRet $\belief_{\text{best}}$\;
\end{algorithm}

\end{minipage}

\section{Bounds}
\label{sec:bounds}

AEMS-SR requires a lower and upper bound $L, U$ for Eq. \ref{eq:upper_belief}, \ref{eq:lower_belief} and \ref{eq:error_approximation}. Those bounds, computed offline, are represented as a set $\Gamma$ of $|\actions|$  $\alpha$-vectors, one for each action $\action$ and denoted as $\alpha_\action$. 
Evaluation of the bounds on a belief $\belief$ is given by $\max_{\alpha \in \Gamma} \tuple{\belief, \alpha}$. 
Typically, the lower bound is derived from Blind policies \citep{hauskrecht1997incremental} that consistently select the same action. For the upper bound, the two main algorithms are QMDP \citep{CassandraIncrementalPrunning} and FIB \citep{Hauskrecht_2000}. %
As the agent retains the option not to request the state, the ability to request the state cannot reduce the expected return but it may potentially increase it. Therefore, ensuring the validity of the upper bounds in POMDP-SRs is crucial.

\smallparagraph{Q-MDP} is constructed by assuming that the uncertainty about the state will disappear after one step and corresponds to solving the underlying MDP. The $\alpha-$vector $\alpha_\action$ is the fixed point of Eq. \ref{eq:qmdp}, and $\alpha_\action\fun{\state}$ corresponds to the Q-Value $Q\fun{\state, \action}$ of the underlying MDP. 
\begin{align}
    \alpha_\action\fun\state = \rewards\fun{\state, \action} + \gamma \sum_{\state' \in \states} \probtransitions\fun{\state'\mid \state, \action} \max_{\alpha_{\action'} \in \Gamma} \alpha_{\action'}\fun{\state'}
    \label{eq:qmdp}
\end{align}

\begin{lemma}
    The Q-MDP upper bound of a standard POMDP $\pomdp$ is an upper bound for the equivalent POMDP $\pomdp'$ of $\pomdpsr$.
\end{lemma}
\begin{proof}
    In the underlying MDP $\mdp'$ of the Equivalent POMDP, the optimal policy will avoid state requests due to the penalty and the full observability. Consequently, the optimal policies of both MDPs only differ by the inclusion of non-request actions, which carry zero reward and thus do not affect the expected return. Hence, for any $\state \in \states$, the value of the optimal policy in $\mdp$ at $\state$ matches the one in $\mdp'$ at $\state^0$. Therefore QMDP is an upper bound for the values of $\pomdpsr$.%
\end{proof}
\smallparagraph{Fast Informed Bound} (FIB) 
considers the partial observability at the next step which provides a tighter upper bound compared to Q-MDP. %
The associated $\alpha$-vectors are constructed by iteratively applying the operator associated to Eq. \ref{eq:fib}.%
\begin{align}
    \alpha_\action\fun\state = \rewards\fun{\state, \action} + \gamma \sum_{\observation \in \observations} \max_{\alpha_{\action'} \in \Gamma} \sum_{\state' \in \states}    P\fun{\state', \observation \mid \state, \action} \alpha_{\action'}\fun{\state'}
    \label{eq:fib}
\end{align}

\begin{lemma}
    The Fast Informed Bound upper bound of a standard POMDP $\pomdp$ is not guaranteed to be an upper bound for $\pomdpsr$'s equivalent POMDP $\pomdp'$.
\end{lemma}

\begin{proof}
    Consider a POMDP $\pomdp$ with two states, a uniform probability transition function, a unique observation, and two actions such that $\rewards\fun{\state_1, \action_1} = \rewards\fun{\state_2, \action_2} = 1$, $\rewards\fun{\state_1, \action_2} = \rewards\fun{\state_2, \action_1} = -1$. 
    Q-MDP returns $(\frac{\gamma}{1 - \gamma} + 1, \frac{\gamma}{1 - \gamma} - 1)$ and $(\frac{\gamma}{1 - \gamma} - 1, \frac{\gamma}{1 - \gamma} + 1)$ as $\alpha$-vectors, which correspond to the uncertainty of the initial state $\pm 1$ and observing to the following states. Conversely, FIB returns $(1, -1)$ and $(-1, 1)$ as $\alpha$-vectors corresponding to the first reward followed by an expected future return of $0$. Let us now consider the POMDP-SR $\pomdpsr=\tuple{\pomdp, \cost}$ and the policy that always request the state to then select the optimal action. This policy has an expected discounted return equal to $\frac{1 - \cost}{1 - \gamma}$. Setting the cost $\cost$ to $0.1$ proves that FIB is not an upper bound for POMDP-SR.  
\end{proof}

\smallparagraph{FIB-SR}~
Adapting FIB to POMDP-SR involves introducing an additional $\alpha$-vector, $\alpha_c$, corresponding to the action of requesting the state. FIB-SR alternates between updating $\alpha$-vectors for environmental actions using Eq. \ref{eq:fib} with $\Gamma=\{\alpha_a, \forall \action \in \actions\} \cup {\alpha_c}$ and updating $\alpha_c$ using Eq. \ref{eq:fib-sr}. This process reflects paying the cost $c$ to observe the state and then selecting the environmental action.
\begin{align}
    \alpha_c\fun\state = - \cost + \max_{\action \in \actions} \alpha_\action\fun\state 
    \label{eq:fib-sr}
\end{align}

\subsection*{Improving the bounds during learning}
In traditional POMDPs, maintaining offline-computed bounds unchanged during online phases is standard practice.
While updating these bounds could enhance the algorithm's efficiency over successive time steps and episodes, this approach is generally not feasible. The primary obstacle is the need to store additional alpha vectors, which would diminish the efficiency of computing bounds for new beliefs and increase memory demands.
Conversely, in POMDP-SRs, corner beliefs present an opportunity to update bounds efficiently. For every corner belief $\state$ in the graph $\graph$ and action $\action$, $U_\graph\fun{\state, \action}$ provides a tighter bound than $\alpha_a\fun{\state}$ computed offline. 
Therefore, by replacing $\alpha_a\fun{\state}$ with the value of $U_\graph\fun{\state, \action}$, we can update the upper bound without requiring additional memory.
The same approach applies to lower bounds, resulting in a practical and efficient solution.

\section{Experiments}

Many existing POMDP benchmarks, like RockSample \citep{hsvi}, feature partial observability that can be permanently eliminated with a single state request, thereby rendering the problem trivial. We evaluate AEMS-SR on Tag, where partial observability is restored at the next timestep, and on RobotDelivery, a new benchmark tailored to POMDP-SR.

\begin{wrapfigure}{r}{0.45\textwidth}
    \centering
    \includegraphics[width=.29\textwidth]{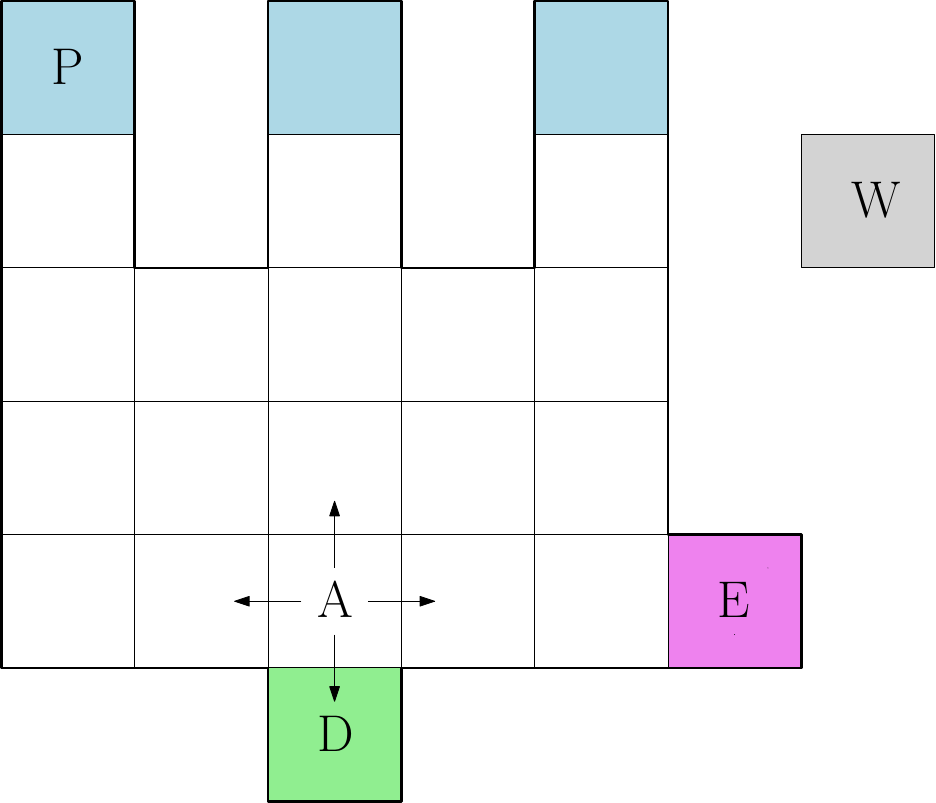}
    \caption{RobotDelivery ($n=3$), %
    A is the agent, P the package, D (green) the delivery location, W (grey) the package waiting area, E (violet) the exit, and the blue tiles are possible package locations.%
    }
    \label{fig:robot_delivery}
\end{wrapfigure}

\smallparagraph{RobotDelivery} 
is a grid-world environment (Fig.~\ref{fig:robot_delivery}) 
featuring a main room (width $3$, length $2n + 1$) with, at the top, $n$ corridors, each two units long and leading to a package pickup point.
At the beginning of each episode, the agent starts at (A) and a package is in one of the pickup-points, with equal probability. Its mission is to collect the package and deliver it to point (D), receiving a reward of $1$ for each successful delivery. After each delivery, there is a probability $e$ that no new packages will spawn. 
Package spawning occurs with probability $1-t$ into the waiting area (W) and with probability $t$ in one of the pickup points. If a package is in the waiting area, it has a probability $t$ of being transferred to a pickup point. The waiting area forces the agent to time its state request as requesting the state when the package is in (W) would force the agent to request it again.  
The agent has four possible actions (up, down, left, right). Except when moving into a package location or delivery location, actions have a failure probability $f$ causing the agent to remain stationary. There are $5$ observations, the first four indicate the number of walls surrounding the agent (0-3), and the fifth occurs when going up to a pickup point with a package or going down into the delivery zone.

\begin{table}
\caption{Experiment results comparing POMCP, AEMS and AEMS-SR with a time limit of 0.1s, 0.5s and 1s on Robot Delivery (with 3, 5 and 7 corridors and cost $c=0.1$) and Tag (cost $c=1$). 
    AEMS and AEMS-SR use the FIB-SR upper bound and \textbf{do not improve the offline bounds during planning}.
We report the mean and standard error for the return and the Error Reduction (ER), and the mean for the Number of Expansions (NE).}

\label{tab:exp_main_sp1}
\resizebox{\linewidth}{!}{%
\begin{tabular}{@{ }c@{  }c|ccc|c@{ }c|c@{ }c|}
\toprule
 &  &  \multicolumn{3}{c|}{Return} & \multicolumn{2}{c|}{Number of Expansions} & \multicolumn{2}{c|}{Error Reduction (\%)} \\
 & T & {POMCP} & {AEMS} & {AEMS-SR} & {AEMS} & {AEMS-SR} & {AEMS} & {AEMS-SR} \\
\midrule
\multirow[c]{3}{*}{RobotDelivery 3} & 0.1 & 0.97$\pm$0.00  & 0.98$\pm$0.00 & \textbf{1.38$\pm$0.01} & 55 & \textbf{2525} & 3.8$\pm$0.0 & \textbf{24.1$\pm$0.4} \\
 & 0.5 & 0.97$\pm$0.00  & 0.98$\pm$0.00 & \textbf{2.49$\pm$0.03} & 111 & \textbf{11150} & 4.4$\pm$0.0 & \textbf{42.4$\pm$0.1} \\
 & 1.0 & 0.97$\pm$0.00  & 0.98$\pm$0.00 & \textbf{2.50$\pm$0.03} & 152 & \textbf{20035} & 4.6$\pm$0.0 & \textbf{43.7$\pm$0.1} \\
\midrule
\multirow[c]{3}{*}{RobotDelivery 5} & 0.1 & 0.95$\pm$0.00  & 0.96$\pm$0.00 & \textbf{1.00$\pm$0.01} & 29 & \textbf{1292} & 3.2$\pm$0.0 & \textbf{7.6$\pm$0.2} \\
 & 0.5 & 0.95$\pm$0.00  & 0.96$\pm$0.00 & \textbf{1.01$\pm$0.01} & 59 & \textbf{6668} & 3.7$\pm$0.0 & \textbf{8.6$\pm$0.2} \\
 & 1.0 & 0.95$\pm$0.00  & 0.96$\pm$0.00 & \textbf{1.45$\pm$0.01} & 80 & \textbf{13347} & 4.0$\pm$0.0 & \textbf{32.3$\pm$0.3} \\
\midrule
\multirow[c]{3}{*}{RobotDelivery 7} & 0.1 & \textbf{0.93$\pm$0.00 } & \textbf{0.93$\pm$0.00} & \textbf{0.93$\pm$0.00} & 17 & \textbf{696} & 2.5$\pm$0.0 & \textbf{6.2$\pm$0.0} \\
 & 0.5 & 0.93$\pm$0.00  & \textbf{0.94$\pm$0.00} & \textbf{0.94$\pm$0.00} & 37 & \textbf{4024} & 3.4$\pm$0.0 & \textbf{6.9$\pm$0.0} \\
 & 1.0 & 0.93$\pm$0.00  & \textbf{0.94$\pm$0.00} & \textbf{0.94$\pm$0.00} & 51 & \textbf{8071} & 3.6$\pm$0.0 & \textbf{7.1$\pm$0.0} \\ 
\midrule
 & 0.1 & -17.43$\pm$0.06  & -6.30$\pm$0.06 & \textbf{-4.66$\pm$0.06} & 20 & \textbf{67} & 54.7$\pm$0.1 & \textbf{58.8$\pm$0.1} \\
Tag  & 0.5 & -17.45$\pm$0.06  & -5.35$\pm$0.08 & \textbf{-4.56$\pm$0.09} & 56 & \textbf{566} & \textbf{67.8$\pm$0.1} & 64.2$\pm$0.1 \\ 
 & 1.0 & -17.47$\pm$0.06  & -5.35$\pm$0.09 & \textbf{-4.50$\pm$0.09} & 79 & \textbf{1124} & \textbf{69.6$\pm$0.1} & 64.9$\pm$0.1  \\
 
\bottomrule
\end{tabular}
}%
\end{table}

\smallparagraph{Tag} is a grid-world environment ($|\states| = 842, |\observations|=30$) where the agent chases a moving prey \citep{PBVI}. The agent observes its own position if not in the same tile as the prey and a special observation otherwise. There are 5 actions, 4 corresponding to cardinal directions, each resulting in a reward of $-1$, and one action to tag the prey yielding $-10$ if not in the same tile and $+10$ otherwise. Successfully tagging the prey terminates the episode. The prey observes both positions, staying in place with a $0.2$ probability, and only moves to increase its distance from the agent. %

\smallparagraph{Setup}
We evaluate AEMS-SR on RobotDelivery environments with 3, 5, and 7 corridors, resulting in state spaces of $133$, $305$, and $541$, respectively, along with Tag. For RobotDelivery, we set $\gamma=0.99$, $f=0.1$, $t=0.8$, and $e=1/3$ to achieve an expected total number of packages as $3$. For Tag, we use $\gamma=0.95$. We tested 0.1s, 0.5s, and 1s as the time per action $T$. We compare against AEMS and POMCP, both running on the Equivalent POMDP, and modified to ensure a fair evaluation by considering the structure of the Equivalent POMDP. AEMS and AEMS-SR both utilize the FIB-SR upper bound. Additional implementation details and experiments, including the Q-MDP upper bound, can be found in the Appendix.
We use R-n and R-n-T to refer to RobotDelivery with n corridors and T seconds of compute time, and T-T for Tag.

\smallparagraph{Metrics} For AEMS and AEMS-SR, in addition of the discounted return, we report the following metrics: 
(NE): Number of Expansions;
(ER): Error Reduction  $1 - \fun{U_\graph\fun\rootBelief - L_\graph\fun\rootBelief}/\fun{U\fun\rootBelief - L\fun\rootBelief}$. We note that the return is not necessarily correlated with the error reduction.

\begin{table}
\caption{Experiment results comparing AEMS and AEMS-SR with a time limit of 0.1s, 0.5s and 1s on Robot Delivery (with 3, 5 and 7 corridors and cost $c=0.1$) and Tag (cost $c=1$). 
    AEMS and AEMS-SR use the FIB-SR upper bound and \textbf{improve the offline bounds during planning}.
We report the mean and standard error for the return and the Error Reduction (ER), and the mean for the Number of Expansions (NE).}

\label{tab:exp_main_sp2}
\center
\resizebox{\linewidth-2cm}{!}{%
\begin{tabular}{@{ }c@{  }c|cc|c@{ }c|c@{ }c|}
\toprule
 &  &  \multicolumn{2}{c|}{Return} & \multicolumn{2}{c|}{Number of Expansions} & \multicolumn{2}{c|}{Error Reduction (\%)} \\
 & T &  {AEMS} & {AEMS-SR} & {AEMS} & {AEMS-SR} & {AEMS} & {AEMS-SR} \\
\midrule
\multirow[c]{3}{*}{RobotDelivery 3} & 0.1 &0.98$\pm$0.00 & \textbf{2.33$\pm$0.02} & 54 & \textbf{2170} & 3.8$\pm$0.0 & \textbf{72.6$\pm$0.5} \\
 & 0.5 & 0.98$\pm$0.00 & \textbf{2.11$\pm$0.01} & 111 & \textbf{8106} & 4.4$\pm$0.0 & \textbf{77.2$\pm$0.5} \\
 & 1.0 & 0.98$\pm$0.00 & \textbf{2.03$\pm$0.01} & 151 & \textbf{13279} & 5.7$\pm$0.2 & \textbf{79.2$\pm$0.5} \\
\midrule
\multirow[c]{3}{*}{RobotDelivery 5} & 0.1 &0.96$\pm$0.00 & \textbf{2.17$\pm$0.02} & 29 & \textbf{1252} & 3.2$\pm$0.0 & \textbf{52.2$\pm$0.4} \\
 & 0.5 &  0.96$\pm$0.00 & \textbf{2.25$\pm$0.02} & 59 & \textbf{5853} & 3.7$\pm$0.0 & \textbf{64.8$\pm$0.6} \\
 & 1.0 &   0.96$\pm$0.00 & \textbf{2.21$\pm$0.02} & 80 & \textbf{10581} & 4.0$\pm$0.0 & \textbf{67.4$\pm$0.6} \\
\midrule
\multirow[c]{3}{*}{RobotDelivery 7} & 0.1 &  0.94$\pm$0.00 & \textbf{1.84$\pm$0.02} & 17 & \textbf{713} & 2.5$\pm$0.0 & \textbf{41.1$\pm$0.5} \\
 & 0.5 &  0.94$\pm$0.00 & \textbf{2.14$\pm$0.02} & 37 & \textbf{3729} & 3.4$\pm$0.0 & \textbf{53.2$\pm$0.5} \\
 & 1.0 &  0.94$\pm$0.00 & \textbf{2.16$\pm$0.02} & 52 & \textbf{7227} & 3.6$\pm$0.0 & \textbf{56.5$\pm$0.5} \\
\midrule
 & 0.1 &  -6.11$\pm$0.06 & \textbf{-4.83$\pm$0.07} & 20 & \textbf{68} & 60.8$\pm$0.1 & \textbf{70.6$\pm$0.1} \\
Tag  & 0.5& -5.41$\pm$0.09 & \textbf{-4.26$\pm$0.09} & 55 & \textbf{571} & 73.7$\pm$0.1 & \textbf{80.8$\pm$0.1} \\
 & 1.0 & -5.23$\pm$0.09 & \textbf{-4.53$\pm$0.09} & 78 & \textbf{1144} & 74.6$\pm$0.1 & \textbf{82.2$\pm$0.1} \\
 
\bottomrule
\end{tabular}
}%
\end{table}

Table \ref{tab:exp_main_sp1} reports the results for POMCP, and for AEMS and AEMS-SR without the improvement of bounds. Table \ref{tab:exp_main_sp2} reports the results for AEMS and AEMS-SR with the improvement of the bounds during planning. 

In RobotDelivery, POMCP consistently yields an average return below 1, opting to exit the room immediately without delivering any packages or making state requests. This poor performance is attributed to the sparse rewards. In Tag, POMCP generally fails to tag the prey 
and obtains an average return inferior to $-17$. 

\smallparagraph{Results without Improving the Bounds}~
In RobotDelivery, AEMS exhibits a strategy similar to POMCP's. In contrast, AEMS-SR achieves an average return of at least 1 in R-3 and R-5 environments, indicating successful package delivery before exiting. Notably, in R-3, with a minimum of 0.5s per step, AEMS-SR's return increases to 2.49, reflecting multiple deliveries. However, in the more complex R-7 scenario, AEMS-SR reverts to an exit-immediately strategy, suggesting potential areas for further enhancement.
In Tag, AEMS performs better than POMCP, as the agent manages to tag the prey, but it is still outperformed by AEMS-SR.

The ER metric, which improves as $T$ grows, aligns with the strategy of expanding the belief deemed to have the greatest impact on reducing the error gap at the root. Interestingly, in Tag, AEMS exhibits a higher ER than AEMS-SR. This is attributed to AEMS episodes being longer, and the initial steps having a very low ER.

To understand the superior performance of AEMS-SR, we can compare its NE against AEMS. 
AEMS-SR conducts up to two orders of magnitude more updates. 
The reason of this difference is that AEMS's belief expansion spawns a new sub-tree for each state in the support (Fig.~\ref{fig:sub1}). This requires computing many belief updates which is time-consuming. AEMS-SR by leveraging the graph structure reuses the already expanded corner beliefs, avoiding unnecessary computations.
This empirically proves the advantage of representing the search space as a graph instead of a tree. %

\smallparagraph{Results with Improving the Bounds}~ 
As detailed in Section~\ref{sec:bounds}, the offline bounds can be easily improved by using the corner beliefs.
For AEMS, however, such refinement has minimal impact on the average return due to the limited number of update steps available for substantial bound improvement.
In contrast, AEMS-SR displays notable performance gains in RobotDelivery. Particularly in the R-7 environment, moving beyond the exit-immediately strategy it delivers packages and obtains an average return over 2 if given at least 0.5s and 1.84 otherwise.  This improvement highlights the efficacy of AEMS-SR combined with online bounds enhancement.
In some instances of AEMS-SR, we observe a significant decrease in NE compared to the non-bounds-improving variant, suggesting that AEMS-SR reaches an $\varepsilon$ solution before the allocated time ends. Additionally, the ER metric, based on the original offline bounds, is higher in scenarios with bounds improvement. Both observations are positive indicators of the algorithm's efficiency.

Overall, the experiments demonstrate AEMS-SR's superior performance over AEMS and POMCP in POMDP-SRs, emphasizing the potential for bounds improvement during the learning phase.

\section{Related Work}

\smallparagraph{Heuristic search}~
Other heuristic search algorithms are notable in the realm of online planning for POMDPs. The approach by \citet{satia1973markovian} employs a branch and bound strategy and utilizes a heuristic similar to AEMS, wherein fringe beliefs are weighted by their likelihood of observation. A key distinction, however, is that all non-dominated actions are deemed equally probable. The BI-POMDP algorithm \citep{washington1997bipomdp} aligns more closely with AO*, and therefore AEMS(-SR), focusing only on fringe nodes accessible with a greedy policy which selects the action that maximizes the upper bound, akin to our Equation \ref{eq:policy_t}. Unlike AEMS, BI-POMDP does not impose additional weighting on the probability of reaching a particular fringe node and instead prioritizes node expansion based on maximizing the error gap. 
In this work, we decided to extend over AEMS because it was shown to be more efficient \citep{OnlinePlanningSurvey}. However, our approach is not limited to AEMS and could be applied to other heuristic search algorithms.

\smallparagraph{State requests in POMDPs}
have seen growing research interest. 
\citet{bellinger2021active} developed the AMRL framework, where agents incur a cost to request \textit{the next state}. This framework, unlike our POMDP-SR, delays state access and doubles the action space instead of separating the two decision steps. 
Their AMRL-Q algorithm, based on Q-learning \citep{watkins1992q}, focuses on state-conditioned policies rather than histories or beliefs, making it sub-optimal.
ACNO-MDPs \citep{ACNO} differ from AMRL by not providing observations without state requests, thus simplifying belief updates. They propose two RL methods: `observe-before-planning', combining initial MDP learning with subsequent POMCP application, and `observe-while-planning', where POMCP or DVRL \citep{igl2018deep} in its deep learning variant, make decisions on state requests and environmental actions.
\citet{act-then-measure} further investigate ACNO-MDPs, focusing on timing state requests through heuristics. While these approaches offer valuable insights, they contrast with our methodology of planning with a pre-known model and striving for an $\varepsilon$-optimal solution. %

\section{Conclusion}
To address environments where the agent can obtain full state information before each action at a cost, we introduce the POMDP with State Requests framework. Within this framework, we present AEMS-SR, a principled algorithm that effectively tackles the exponential growth challenge in POMDP-SR tree-based search by employing a cyclic graph structure.
Our theoretical analysis proved that AEMS-SR is complete and $\varepsilon$-optimal.
Empirical evaluation in RobotDelivery — a novel benchmark designed for POMDP-SR — and Tag 
demonstrates AEMS-SR's superior performance compared to established algorithms, AEMS and POMCP, in POMDP-SR settings. 
In future work, we aim to develop policies $\hat{\policy}_\graph$ tuned to the specificity of POMDP-SR, and we plan to investigate the potential application of AEMS-Loop to other subclasses of POMDPs.%

    \subsection*{Acknowledgments}
Rapha\"el Avalos is supported by the Research Foundation – Flanders (FWO), under grant number 11F5721N, and under 
grant number V418823N for the research visit at TU Delft. This research was supported by funding from the Flemish Government under the ``Onderzoeksprogramma Artifici\"{e}le Intelligentie (AI) Vlaanderen'' program.
We would like to thank all the members of TU Delft's Interactive Intelligence group for making this research visit 
enjoyable and fruitful.
Finally, we would like to thank Willem R\"opke for his help in proofreading the paper.

\bibliography{references}
\bibliographystyle{rlc}
\newpage
\appendix

\section{Equivalent POMDP}
In this section, we explain the technicalities of the transformation of a POMDP-SR $\pomdpsr = \pomdpsrtuple$ into an equivalent POMDP $\pomdp' = \langle \states', \observations', \actions', \actionfn', \probtransitions', \observationfn', \rewards',  \discount' \rangle$  with variable action space and with $\probtransitions', \observationfn', \rewards'$ defined only over legal actions. 
\begin{itemize}
    \item The state space $\states'\equiv \{0, 1\} \times \states$ augments the original state space with a binary indicator $i$, which equals 0 when the agent needs to decide whether to request the state and 1 for selecting the environmental action. To ease notations, we add the binary indicator in sup-script $\state^i$.
    \item The observations space $\observations' = \observations\cup\states\cup\{\observation^*\}$; $\observation^*$ is a special observation associated with not requesting the state.
    \item The action space $\actions' \equiv \{\notrequest, \request\} \cup \actions$ includes additional actions: $\request$ for requesting the state, and $\notrequest$ not to request it. The set of legal actions is returned by $\actionfn$ which is defined as follows $\actionfn\fun{\state^0} = \{\notrequest, \request\}$, $\actionfn\fun{\state^1} = \actions$. We note that, since at each time step the states in the support of the belief have always the same binary indicator $i$, $\actionfn$ can be extended to beliefs.
    \item The transition function $\probtransitions'$ is defined for all $\state, \state' \in \states$ as follows, $\probtransitions'\fun{\state'^0 | \state^1, a} = \probtransitions\fun{\state' | \state, a}$ and $\probtransitions'\fun{\state'^1 | \state^0, a} = 1_{\state}\fun{\state'}$, $\probtransitions'\fun{\state'^i | \state^i, a} = 0$, with $1$ the indicator function,
    \item The observation function $\observationfn': \states \times \actions \rightarrow \Delta_\observation$ is defined for legal actions as follows, $\observationfn'\fun{\state'^0, a} = \observationfn\fun{\state', a}$,  $\observationfn'\fun{\state'^1, \request}$ is a dirac distribution centered in $\state'$, and $\observationfn'\fun{\state'^1, \notrequest}$is a dirac distribution centered in $\observation^*$
    \item The reward function $\rewards'$ is defined for legal actions as follows: $\rewards'\fun{\state^0, a} = - 1_{\request}\fun{a}  c / \sqrt{\gamma}$, $\rewards'\fun{\state^1, a} = \rewards\fun{\state, a}$
    \item $\gamma'=\sqrt{\gamma}$
\end{itemize}

\section{Example of Optimal Action Divergence between MDP, POMDP and POMDP-SR}

\begin{figure*}[th]
    \centering
    \includegraphics[width=.9\linewidth]{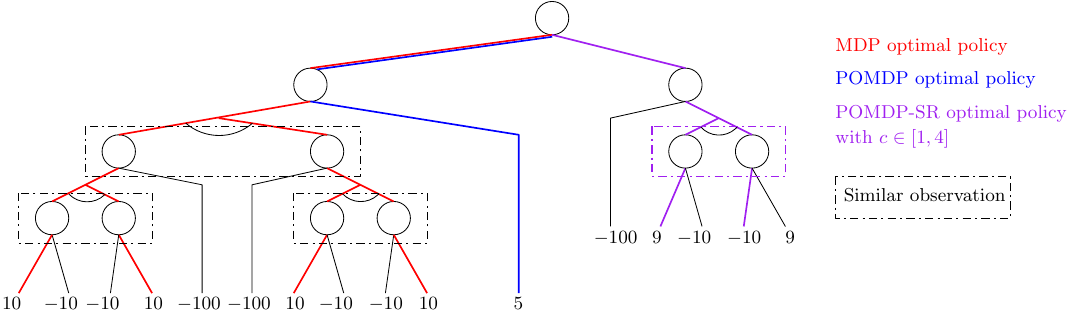}
    \caption{Tree representing an environment where the MDP and POMDP optimal action are the same but if the agent can request the state for a cost $1 \leq \cost \leq 4$ the optimal action changes. Circles represent states, actions are left $a_1$ and right $a_2$,  }
    \label{fig:pomdrs_diff_action}
\end{figure*}

As discussed in the Framework section, the optimal action in a POMDP-SR scenario may deviate from that in an MDP or POMDP, even when their optimal actions are aligned. This divergence is attributed to the capability of requesting state information in future steps. To illustrate this concept, we present a toy environment, depicted in Figure \ref{fig:pomdrs_diff_action}, where such a divergence occurs.

In our example environment, states producing the same observations are enclosed within dashed boxes. Initially, at state $\state_0$, both MDP and POMDP strategies suggest executing action 1, which leads to expected returns of 10 and 5, respectively. However, in the POMDP-SR, the calculation of Q-values yields $\max(5, 10 - 2c)$ for action 1 and $9 - c$ for action 2. Consequently, action 2 emerges as the optimal choice when the cost values $c$ lie within the range of $[1, 4]$. This example demonstrates how POMDP-SR compels the agent to operate with a forward-looking approach, considering potential future state requests. This adds a layer of complexity to the decision-making process, differing significantly from situations in classic POMDPs where state requests are either limited to the current timestep or entirely absent.

\section{Notations}

We starts by restating some definitions and notations.%
\begin{itemize}
    \item We denote trees as $\tree$ and graphs as $\graph$.
    \item $\fringe\fun{\graph}$ is the set of nodes in graph $\graph$ that does not have any children. We use the same notation for trees $\fringe\fun{\tree}$. 
    \item Corner beliefs corresponds to beliefs where one of the state has probability $1$. We use the notation $s$ to refer both to the state and the associated corner belief.
    \item $\pathSet_\graph\fun{\rootBelief, \belief}$ is the set of paths in the graph $\graph$ that starts on $\rootBelief$ and finishes on $\belief$.
    \item Paths $\history$ are sequences of beliefs, action and observations $(\belief_{i}, \action_i, \observation_{i+1})_{i < T}$ with $T=d(\history)$ being its length. We write paths that start in $\belief_1$ and ends in $\belief_2$ as $\history_{\belief_1}^{\belief_2}$.
    \item $P\fun{\history | \rootBelief, \pi} = \Pi_{i=0}^{T - 1} P\fun{\observation_{i+1} | \belief_i, \action_i} \pi\fun{\action_i | \belief_i}$ corresponds to the probability of observing the path $\history$ while starting at the root belief $\rootBelief$ and following the policy $\pi$.
    \item $\cumProb^\graph_{\pi}\fun{\rootBelief, \belief}= \sum_{\history \in \pathSet_\graph\fun{\rootBelief, \belief}} \gamma^{d\fun{\history}} P\fun{\history | \rootBelief, \policy}$ corresponds to the sum over all possible paths between the root belief $\rootBelief$ and a belief $\belief$ of the probability of observing the path discounted by its length.
    \item For any belief $\belief \in \beliefs$ and lower bound $L$, we defined the error gap as $\gap{\belief} = V^*\fun{\belief} - L\fun{\belief}$
    \item For any belief $\belief \in \beliefs$, lower bound $L$ and upper bound $U$, we defined the approximate error gap as $\gapApprox{}{\belief} = U\fun{\belief} - L\fun{\belief} \geq e\fun{\belief}$
\end{itemize}

\section{Proofs}

\begin{theorem}
    In any rooted graph $\graph$ with root $\rootBelief$ where values are computed according to Equation \ref{eq:lower_belief} using a lower bound value function L, bounded bellow, with error $e(\belief) = V^*(\belief) - L(\belief)$, the error on the root belief state is bounded by: 
    $e_\graph(\rootBelief) = V^*(\rootBelief) - L_\graph(\rootBelief) \leq \sum_{\belief \in \fringe\fun{\graph}} \cumProb_{\pi^*}\fun{\rootBelief, \belief} e(\belief)$. where $e(\belief) = V^*(\belief) - L(\belief)$.
    \label{th:th1-appendix}
\end{theorem}

\begin{proof}
    Consider an arbitrary node $\belief \in \graph \setminus \fringe\fun{\graph}$ in a graph $\graph$ that is not a fringe, and $\action^*_\belief = \arg\max_\action Q^*(\belief, \action)$ the optimal action. By definition $\gamma \in [0, 1)$. If $\gamma = 0$, then $L\fun{\belief} = V^*\fun{\belief} = \reward\fun{\belief, \action^*_\belief}$ and $e_\graph\fun{\belief} = 0$ which conclude the proof. Therefore let us focus on $\gamma \in (0, 1)$. 
    By definition of $L_\graph$ we have $L_\graph\fun{\belief, \action^*_\belief} \leq  L_\graph(\belief)$. 
    
    \begin{align*}
        e_\graph\fun{\belief} &= V^*(\rootBelief) - L_\graph(\rootBelief) \\
        &\leq V^*(\rootBelief) - L_\graph(\rootBelief, \action^*_\belief) \\
        &\leq \reward\fun{\belief, \action^*_\belief} + \gamma \sum_{\observation \in \observations} P\fun{\observation | \belief, \action^*_\belief} V^*\fun{\tau\fun{\belief, \action^*_\belief, \observation}} - \left(\reward\fun{\belief, \action^*_\belief} + \gamma \sum_{\observation \in \observations} P\fun{\observation | \belief, \action^*_\belief} L_\graph\fun{\tau\fun{\belief, \action^*_\belief, \observation}}\right) \\
        &\leq \gamma \sum_{\observation \in \observations} P\fun{\observation | \belief, \action^*_\belief} e_\graph\fun{\tau\fun{\belief, \action^*_\belief, \observation}}
    \end{align*}
    
    This result in the following inequality:
    \begin{align}
        e_\graph(\belief) \leq \begin{cases}
            e(\belief) & \text{if $\belief \in \fringe\fun\graph$} \\
            \gamma \sum_{\observation \in \observations} P\fun{\observation | \belief, \action^*_\belief} e_\graph\fun{\tau\fun{\belief, \action^*_\belief, \observation}} & \text{otherwise}
        \end{cases}
        \label{eq:appendix_graph_error_ineq}
    \end{align}
    Let us now consider $n \in \mathbb{N}^*$, and unroll the rooted graph $\graph$ into a tree $\tree_n$ with root $\rootBelief$ and with maximum depth $n$. We define the following elements:
    \begin{itemize}
        \item Similar to our definition of $\pathSet$, we define, for all $\belief' \in \graph$, $\pathSet_n\fun{\rootBelief, \belief'}$ the set of paths starting from $\rootBelief$ and finishing in any of the replicas of $\belief'$ in $\tree_n$. It is important to note that the length of the paths in $\pathSet_n$ are bounded by $n$. We have $\lim_{n \rightarrow +\infty} \pathSet_n \fun{\rootBelief, \belief'} = \pathSet  \fun{\rootBelief, \belief}$ the (possibly infinite) set of paths between $\rootBelief$ and $\belief'$ in $G$. 
        \item For any $\belief' \in \graph$ and any policy $\pi$, $\cumProb^n_{\pi}\fun{\rootBelief, \belief'}= \sum_{\history \in \pathSet_n\fun{\rootBelief, \belief'}} \gamma^{d\fun{\history}} \probtransitions\fun{\history | \rootBelief, \policy }$
        \item $\fringe_n$ as the fringes nodes of $\tree_n$
        \item $\fringe^\graph_n$ is the set of nodes of $\fringe_n$ that are replicas of a node in $\fringe\fun{\graph}$
        \item $\bar{\fringe^\graph_n} = \fringe_n \setminus \fringe^\graph_n$
        \item $e_n = e_{\tree_n}$
    \end{itemize}
    
    We note that Equation \ref{eq:appendix_graph_error_ineq} hold for any graph including $\tree_n$. Therefore, by solving the recurrence in $\tree_n$ for $\rootBelief$ we obtain (as in the proof of Theorem 1 of AEMS): 
    \begin{align*}
        e_n\fun{\rootBelief} &\leq \sum_{\belief \in \fringe_n} \gamma^{d\fun{\rootBelief, \belief}} P\fun{\history_{\rootBelief}^\belief | \rootBelief, \pi^*} e\fun{\belief} \\
        &\leq \sum_{\belief \in \fringe^\graph_n} \gamma^{d\fun{\rootBelief, \belief}} P\fun{\history_{\rootBelief}^\belief | \rootBelief, \pi^*} e\fun{\belief} + \sum_{\belief \in\bar{\fringe}^\graph_n} \gamma^{d\fun{\rootBelief, \belief}} P\fun{\history_{\rootBelief}^\belief | \rootBelief, \pi^*} e\fun{\belief} \\
        &\leq \sum_{\belief \in \fringe\fun{\graph}} \cumProb^n_{\pi^*}\fun{\rootBelief, \belief} e\fun{\belief} + \sum_{\belief \in\bar{\fringe}^\graph_n} \gamma^{d\fun{\rootBelief, \belief}} P\fun{\history_{\rootBelief}^\belief | \rootBelief, \pi^*} e\fun{\belief}
    \end{align*}
    
    We note that for all $\belief \in \bar{\fringe}^\graph_n$ the associated history $\history$ in $\tree_n$ is of length $n$. Indeed, if the history size was shorter than $n$, $\belief$ would also be a fringe in $\graph$ which is impossible by definition of $\bar{\fringe}^\graph_n$. It follows that:
    
    \begin{align*}
        e_n\fun{\rootBelief} &\leq \sum_{\belief \in \fringe\fun{\graph}} \cumProb^n_{\pi^*}\fun{\rootBelief, \belief} e\fun{\belief} + \gamma^n  \sum_{\belief \in\bar{\fringe}^\graph_n} P\fun{\history_{\rootBelief}^\belief | \rootBelief, \pi^*} e\fun{\belief} \\
        &\leq \sum_{\belief \in \fringe\fun{\graph}} \cumProb^n_{\pi^*}\fun{\rootBelief, \belief} e\fun{\belief} + \gamma^n \sup_{\belief'} e\fun{\belief'} \tag{the $\sup$ exists because $V^*$ is bounded above and $L$ bellow}
    \end{align*}
    
    In the limit, when $n \rightarrow + \infty$ we have $e_n \rightarrow e_\graph$, $\cumProb^n \rightarrow \cumProb$, and $\gamma^n \rightarrow 0$ as $\gamma \in (0, 1)$, leading to

    \begin{align*}
        e_\graph\fun{\rootBelief} \leq \sum_{\belief \in \fringe\fun{\graph}} \cumProb_{\pi^*}\fun{\rootBelief, \belief} e(\belief)
    \end{align*}
\end{proof}

\begin{definition}
We define the approximate error contribution of a fringe node $\belief \in \fringe\fun{\graph}$ on the value at the root $\rootBelief$ as 
\begin{align*}
    E\fun{\belief, \rootBelief, \graph} = \cumProb_{\hat{\pi}_\graph}\fun{\rootBelief, \belief} \gapApprox{}{\belief}
\end{align*}
\end{definition}

\begin{lemma}
    In any graph $\graph$, the approximate error contribution $E\fun{\belief, \rootBelief, \graph}$ of a belief node $\belief$ is bounded by $E\fun{\belief, \rootBelief, \graph} \leq \gamma^{d_\belief} \sup_{\belief'} \gapApprox{}{\belief'}$ with $d_\belief=\min_{\history \in \pathSet_\graph\fun{\rootBelief, \belief}} d\fun{\history}$.
    \label{lemma:bounded_contribution}
\end{lemma}

\begin{proof}
    \begin{align*}
        E\fun{\belief, \rootBelief, \graph} &= \cumProb_{\hat{\pi}_\graph}\fun{\rootBelief, \belief} \gapApprox{}{\belief} \\
        &= \sum_{\history \in \pathSet\fun{\rootBelief, \belief}} \gamma^{d\fun{\history}} \probtransitions\fun{\history | \rootBelief, \hat{\policy}_\graph} \gapApprox{}{\belief} \\
        &\leq \gamma^{d_\belief} \sum_{\history \in \pathSet\fun{\rootBelief, \belief}} \probtransitions\fun{\history | \rootBelief, \hat{\policy}_\graph} \gapApprox{}{\belief} \\
        &\leq \gamma^{d_\belief} \sup_{\belief'} \gapApprox{}{\belief'} \sum_{\history \in \pathSet\fun{\rootBelief, \belief}} \probtransitions\fun{\history | \rootBelief, \hat{\policy}_\graph}  \\
        &\leq \gamma^{d_\belief} \sup_{\belief'} \gapApprox{}{\belief'}
    \end{align*}
\end{proof}

\begin{definition}
    We define the set of accessible fringe nodes of a graph $\graph$ under $\hat{\pi}_\graph$ as $\Hat{\beta}\fun{\graph} = \{ \belief | \belief \in \fringe\fun{\graph} \text{ and } \exists \history \in \pathSet_\graph\fun{\rootBelief, \belief}, P\fun{\history | \rootBelief, \hat{\pi}_\graph} > 0 \} $.
    And the set of possible histories $\hat{\zeta}\fun{\rootBelief,\graph} = \{ \history | P\fun{\history | \rootBelief, \hat{\pi}_\graph} > 0 \}$.
    We define, for any $\belief \in \graph$, the set of possible histories between the root belief $\belief_0$ and $\belief$ as $\pathSet_\graph^p\fun{\rootBelief, \belief} = \{ \history_{\rootBelief}^\belief |  \history_{\rootBelief}^\belief \in \pathSet_\graph\fun{\rootBelief, \belief} \bigcap \hat{\zeta}\fun{\rootBelief,\graph} \}$%
\end{definition}

\begin{definition}
    For all histories $\history=(\belief_i, \action_i, \observation_{i+1})_{i < T}$, where $T=d(\history)$ is the length of the history, we define the observation probability $P\fun{\history | \rootBelief} = \Pi_{i=0}^{T - 1} P\fun{\observation_{i+1} | \belief_i, \action_i}$    
\end{definition}

\begin{lemma}
    Given $U$ bounded above and $L$ bounded bellow such as $U\fun{\belief} \geq V^*\fun{\belief} \geq L\fun{\belief}$, and $\gapApprox{}{\belief} = U\fun{\belief} - L\fun{\belief}$ for all $\belief \in \beliefs$, then for any graph $\graph$, $\varepsilon > 0$ and $D \in \mathbb{N}^*$ such that $\gamma^D \sup_\belief \gapApprox{}{\belief} \leq \varepsilon$, if for all $\belief \in \hat{\beta}\fun{\graph}$ and for all $\history \in \pathSet_\graph^p\fun{\rootBelief, \belief}$, %
    either $d\fun{\history} > D$ or there exists an ancestor $\belief' \in \history$ such that $\gapApprox{\graph}{\belief'} \leq \varepsilon$, then $\gapApprox{\graph}{\rootBelief} \leq \varepsilon$. 
    \label{lemma:bounded_root}
\end{lemma}

\begin{proof}
    For any graph $\graph$, and any belief $\belief$ that is not a fringe belief node $\belief \in \graph \setminus \fringe\fun{\graph}$. 
    We define as $\hat{\action}_\belief^\graph = \arg\max_{\action \in \actions} U_\graph\fun{\belief, \action}$. 
    
    \begin{align*}
        \gapApprox{\graph}{\belief} &= U_\graph\fun{\belief} - L_\graph\fun{\belief}\\
        &\leq  U_\graph\fun{\belief, \hat{\action}_\belief^\graph} - L_\graph\fun{\belief, \hat{\action}_\belief^\graph} \\
        &\leq \reward\fun{\belief, \hat{\action}_\belief^\graph} + \gamma \sum_{\observation \in \observations} P\fun{\observation | \belief, \hat{\action}_\belief^\graph} U_\graph\fun{\tau\fun{\belief, \hat{\action}_\belief^\graph, \observation}} - \left(\reward\fun{\belief, \hat{\action}_\belief^\graph} + \gamma \sum_{\observation \in \observations} P\fun{\observation | \belief, \hat{\action}_\belief^\graph} L_\graph\fun{\tau\fun{\belief, \hat{\action}_\belief^\graph, \observation}})\right) \\
        &\leq \gamma \sum_{\observation \in \observations} P\fun{\observation | \belief, \hat{\action}_\belief^\graph} \gapApprox{\graph}{\tau\fun{\belief, \hat{\action}_\belief^\graph, \observation}}
    \end{align*}
    
    We obtain the following upper bound for $\belief \in \graph$ on $\gapApprox{\graph}{\belief}$:
    \begin{align}
    \gapApprox{\graph}{\belief} \leq \begin{cases}
        \gapApprox{}{\belief} & \text{if $\belief \in \fringe\fun\graph$} \\
        \varepsilon & \text{if $\gapApprox{\graph}{\belief} \leq \varepsilon$} \\
        \gamma \sum_{\observation \in \observations} P\fun{\observation | \belief, \action^*_\belief} \gapApprox{\graph}{\tau\fun{\belief, \hat{\action}^\graph_\belief, \observation}} & \text{otherwise}
    \end{cases}
    \end{align}

    We define : 
    \begin{itemize}
        \item $\gapApprox{\graph}{\history_{\rootBelief}^\belief} = \gapApprox{\graph}{\belief}$
        \item $A(\graph)$ is the set of possible histories $\history_{\rootBelief}^{\belief} \in \hat{\zeta}\fun{\rootBelief,\graph}$ of length $d\fun{\history_{\rootBelief}^{\belief}} < D$ such that $\gapApprox{\graph}{\belief} \leq \varepsilon$.%
        \item $B(\graph)$ is the set of possible histories $\history_{\rootBelief}^{\belief} \in \hat{\zeta}\fun{\rootBelief,\graph}$ of length $d\fun{\history_{\rootBelief}^{\belief}} < D$ such that $\belief \in \fringe\fun{\graph}$, %
        and that for all intermediary $\belief' \in \history_{\rootBelief}^\belief$, the partial history $\history_{\rootBelief}^{\belief'} \not\in A\fun{\graph}$.
        \item $C(\graph)$ is the set of all possible histories $\history_{\rootBelief}^{\belief} \in \hat{\zeta}\fun{\rootBelief,\graph}$ of size at least $D$, %
        that do not belong to $B(\graph)$ and for all intermediary $\belief' \in \history_{\rootBelief}^\belief$,  the partial history $\history_{\rootBelief}^{\belief'} \notin A\fun{\graph}$.
    \end{itemize}

    As for all $\belief \in \hat{\beta}\fun{\graph}$ and for all $\history \in \pathSet_\graph^p\fun{\rootBelief, \belief}$ %
    either $d\fun{\history} > D$ or there exists an ancestor $\belief' \in \history$ such that $\gapApprox{\graph}{\belief'} \leq \varepsilon$, $B\fun{\graph}$ is empty.

    By unfolding the recurrence above we obtain:
    {\allowdisplaybreaks
    \begin{align*}
        \gapApprox{\graph}{\rootBelief} 
        &= \sum_{\history_{\rootBelief}^\belief \in A\fun{\graph}} \gamma^{d\fun{\history_{\rootBelief}^\belief}} P\fun{\history_{\rootBelief}^\belief | \rootBelief} \gapApprox{\graph}{\belief} + \sum_{\history_{\rootBelief}^\belief \in C\fun{\graph}} \gamma^{d\fun{\history_{\rootBelief}^\belief}} P\fun{\history_{\rootBelief}^\belief | \rootBelief} \gapApprox{\graph}{\belief}\\
        &\leq \varepsilon \sum_{\history_{\rootBelief}^\belief \in A\fun{\graph}} P\fun{\history_{\rootBelief}^\belief | \rootBelief} + \sum_{\history_{\rootBelief}^\belief \in C\fun{\graph}} \gamma^{d\fun{\history_{\rootBelief}^\belief}} P\fun{\history_{\rootBelief}^\belief | \rootBelief} \gapApprox{\graph}{\belief} \\
        &\leq \varepsilon \sum_{\history_{\rootBelief}^\belief \in A\fun{\graph}} P\fun{\history_{\rootBelief}^\belief | \rootBelief} + \gamma^D \sup_{\belief} \gapApprox{}{\belief} \sum_{\history_{\rootBelief}^\belief \in C\fun{\graph}} P\fun{\history_{\rootBelief}^\belief | \rootBelief} \\
        &\leq \varepsilon \sum_{\history_{\rootBelief}^\belief \in A\fun{\graph}} P\fun{\history_{\rootBelief}^\belief | \rootBelief} + \varepsilon \sum_{\history_{\rootBelief}^\belief \in C\fun{\graph}} P\fun{\history_{\rootBelief}^\belief | \rootBelief} \\
        &\leq \varepsilon \sum_{\history_{\rootBelief}^\belief \in A\fun{\graph} \cup C\fun{\graph}} P\fun{\history_{\rootBelief}^\belief | \rootBelief} \\
        &\leq \varepsilon
    \end{align*} }   
\end{proof}

\begin{theorem}
    Given $U$ bounded above and $L$ bounded bellow such as $U\fun{\belief} \geq V^*\fun{\belief} \geq L\fun{\belief}$, and $\gapApprox{}{\belief} = U\fun{\belief} - L\fun{\belief}$ for all $\belief \in \beliefs$, if $\gamma \in [0, 1)$ and $\inf_{\belief, \graph | \gapApprox{\graph}{\belief} > \varepsilon} \hat{\pi}_\graph\fun{\belief, \hat{\action}_\belief^\graph} > 0$ for $\hat{\action}_\belief^\graph = \arg\max_{\action \in \actions} U_\graph\fun{\belief, \action}$, then the AEMS-Loop algorithm using heuristic $\Tilde{\belief}\fun{\graph}$ is complete and $\varepsilon-$ optimal.
    \label{th:th2_appendix}
\end{theorem}

\begin{proof}
    Consider an arbitrary $\varepsilon>0$ and the current root belief $\rootBelief$. If $\gamma = 0$, then after one expansion $\gapApprox{\graph}{\rootBelief} = 0 $ since $U_\graph\fun{\rootBelief} = L_\graph\fun{\rootBelief} = \max_{\action \in \actions} \reward\fun{\rootBelief, \action}$. And therefore AEMS-Loop is complete and $\varepsilon-$optimal.\\
    
    Lets focus on $\gamma \in (0, 1)$. Because $U$ is bounded above and $L$ bellow, $\sup_\belief \gapApprox{}{\belief}$ exists and there exists $D \in \mathbb{N}$ such that $\gamma^D \sup_\belief \gapApprox{}{\belief} < \varepsilon$. 
    We define the following elements:
    \begin{itemize}
        \item $\mathcal{A}^\graph\fun{\history_{\rootBelief}^\belief}$ the set of ancestors beliefs of $\belief$ in the history $\history_{\rootBelief}^\belief$
        \item $\mathcal{A}^\graph\fun{\belief} = \bigcup_{\history_{\rootBelief}^\belief \in \pathSet^p_\graph\fun{\belief_0, \belief}} \mathcal{A}^\graph\fun{\history_{\rootBelief}^\belief}$ the set of ancestors beliefs of $\belief$ across possible histories. %
        \item $\hat{e}_\graph^{min}\fun{A} = \min_{\belief \in A} \gapApprox{\graph}{\belief}$ for any finite set of beliefs $A$.
        \item $\graph_\belief=\{ \graph | \graph \text{ is finite}, \belief\in\hat{\beta}\fun{\rootBelief, \graph}, \max_{\history \in \pathSet^p_\graph\fun{\belief_0, \belief}}  \hat{e}^{min}_\graph\fun{\mathcal{A}^\graph\fun{\history}} > \varepsilon \}$ Intuitively, $\graph_\belief$ is the set of finite graphs $\graph'$ with root $\rootBelief$ for which $\belief$ is an accessible fringe node, i.e. that can be attained with non zero probability under the policy $\hat{\pi}_{\graph'}$, and for which there exist an history $\history_{\rootBelief}^\belief$ such that all the ancestors beliefs $\belief'$ in that history have an approximate error gap $\gapApprox{\graph'}{\belief'} > \varepsilon$. The existence of the max relies on the graph being finite. 
        \item $\mathcal{B} = \{ \belief | \gapApprox{}{\belief} \inf_{\graph \in \graph_\belief} \sum_{\history \in \pathSet_\graph\fun{\rootBelief, \belief} | d\fun{\history} \leq D} P\fun{\history_{\rootBelief}^\belief | \rootBelief, \hat{\pi}_\graph} > 0 \}$
    \end{itemize}

    The assumption  $\inf_{\belief, \graph | \gapApprox{\graph}{\belief} > \varepsilon} \hat{\pi}_\graph\fun{\belief, \hat{\action}_\belief^\graph} > 0$ ensures that $\mathcal{B}$ contains all the beliefs states $\belief$ within depth $D$ such that (i) $\gapApprox{}{\belief} > 0$, (ii) there exists a finite graph $\graph$ where $\belief \in \hat{\beta}\fun{\rootBelief, \graph}$ and for which there exists an history $\history_{\rootBelief}^\belief \in \pathSet_\graph^p\fun{\rootBelief, \belief}$ %
    such that all ancestors $\belief' \in \mathcal{A}\fun{\history_{\rootBelief}^\belief}$ have $\gapApprox{\graph}{\belief'} > \varepsilon$. 

    As there are only a finite number of beliefs for which there exists an history of size smaller than $D$, $\mathcal{B}$ is finite. This allows us to define $E_{min} = \min_{\belief \in \mathcal{B}} \gapApprox{}{\belief} \inf_{\graph \in \graph_\belief} \sum_{\history_{\rootBelief}^\belief \in \pathSet_\graph\fun{\rootBelief, \belief}} \gamma^{d\fun{\history_{\rootBelief}^\belief}} P\fun{\history_{\rootBelief}^\belief | \rootBelief, \hat{\pi}_\graph}$.     

    By construction $E_{min} > 0$. We also know that for any graph $\graph$, all beliefs $\belief \in \mathcal{B} \cap \hat{\beta}\fun{\rootBelief, \graph}$ have an approximate error contribution $E\fun{\belief, \rootBelief, \graph} \geq E_{min}$ 
    \begin{align*}
        E\fun{\belief, \rootBelief, \graph} 
        =  \cumProb_{\hat{\pi}_\graph}\fun{\rootBelief, \belief} \gapApprox{}{\belief} 
        = \sum_{\history_{\rootBelief}^\belief \in \pathSet\fun{\rootBelief, \belief}} \gamma^{d\fun{\history_{\rootBelief}^\belief}} P\fun{\history_{\rootBelief}^\belief | \rootBelief, \hat{\policy}_\graph} \gapApprox{}{\belief} 
         \geq E_{\min}
    \end{align*}
    As $\gamma \in (0,1)$ and $ E_{min} >0 $, there exist $D' \in \mathbb{N}^+$ such that $\gamma^{D'} \sup_{\belief} \gapApprox{}{\belief} < E_{min}$. Therefore, we know from Lemma \ref{lemma:bounded_contribution} that AEMS-Loop cannot expand any node of depth $D'$ or more before expanding a graph $\graph$ where $\mathcal{B} \cap \hat{\beta}\fun{\rootBelief, \graph} = \emptyset$.

    As there exist a finite number of belief nodes for which an history starting from $\rootBelief$ of length at most $D'$ exists, it is clear that AEMS-Loop will reach such a graph $\graph$ after a finite number of expansions. 

    Since, for this graph $\graph$, $\mathcal{B} \cap \hat{\beta}\fun{\rootBelief, \graph} = \emptyset$ we have that for all beliefs $\belief \in \hat{\beta}\fun{\rootBelief, \graph}$ the possible histories $\history_{\rootBelief}^\belief \in \pathSet^p_\graph\fun{\rootBelief, \belief}$ with length $d\fun{\history_{\rootBelief}^\belief} \leq D$ have $\hat{e}^{min}_\graph\fun{\mathcal{A}^\graph\fun{\history_{\rootBelief}^\belief}} < \varepsilon$. 
    Therefore, Lemma \ref{lemma:bounded_root} ensures that $\gapApprox{\graph}{\rootBelief} < \varepsilon$ and consequently AEMS-Loop will terminate with an $\varepsilon$-optimal solution in a finite number of expansions as $\hat{e}_\graph$ is an upper bound of $e_\graph$.
\end{proof}

\section{Algorithm to compute $\cumProbDirect$ and $\cumProbDirect_{\rootBelief}$}

Algorithm \ref{algo:getbelieftoexpand_gwalk} presents the pseudo code to compute $\cumProbDirect$ and $\cumProbDirect_{\rootBelief}$. The algorithm recursively traverse the graph by following $\hat{\policy}_\graph$ and by maintaining two sets, one for the visited corner beliefs and one for the fringe beliefs.
\begin{algorithm}
\DontPrintSemicolon
\caption{GWalk: computing $\cumProbDirect$ and $\cumProbDirect_{\rootBelief}$ }\label{alg:gwalk}
\SetKwInOut{Input}{input}
\Input{belief $\belief$, visitedStates $\mathcal{V}$, stateMatrix $M$, $\bar{p}$, fringeNodes $\hat{\fringe}$, Graph $\graph$, root belief $\rootBelief$}
\If{$\belief \in \fringe\fun{\graph}$}{
    $\hat{\fringe} = \hat{\fringe} \cup \{\belief\}$\;
}
\ElseIf{$\pi(\belief) = \request$}{
    \For{$\state \in \support\fun{\belief}$}{
        \If{$\origin\fun{\belief} = \rootBelief$}{
            $\bar{p}$[$\state$] += $\belief[\state] * \cumProbDirect(\origin(\belief), \belief$)\;
        }
        \Else{
            $M_{\origin(\belief), \state} += \belief[\state] * \cumProbDirect(\origin(\belief), \belief$)\;
        }
        \If{$\state \notin \mathcal{V}$}{
            $\mathcal{V} = \mathcal{V} \cup \{\state\}$\;
            \For{$\observation \in \observationfn(\state, \pi(\state))$}{
                $\mathcal{V}, M, \bar{p}, \hat{\fringe} \leftarrow$ \texttt{GWalk}($\tau\fun{\state, \pi(\state), \observation}, \mathcal{V}, M, \bar{p}, \hat{\fringe}$)\;
            }
        }
    }
}
\Else{
    \For{$\observation \in \observationfn(\belief, \pi(\belief))$}{
        $\mathcal{V}, M, \bar{p}, \hat{\fringe} \leftarrow$  \texttt{GWalk}($\tau\fun{\belief, \pi(\belief), \observation}, \mathcal{V}, M, \bar{p}, \hat{\fringe}$)\;
    }
}
\Return{$\mathcal{V}, M, \bar{p}, \hat{\fringe}$}
\end{algorithm}

\section{Experiment Supplementary Details}
\label{app:exp_details}

\smallparagraph{POMCP} Our implementation of POMCP is an adaptation of the one provided by \url{https://github.com/Svalorzen/AI-Toolbox} to handle the variable action space of the extended POMDP. This allows us to avoid resorting to deterrent penalties for illegal actions, which can hinder learning.

\smallparagraph{AEMS} Our implementation of AEMS is also tailored for the POMDP-SR structure. This prevents the need to double the state space, as done in the extended POMDP formulation, ensuring a fair evaluation as doubling the state space would slow the belief update computation.

\smallparagraph{Tag} The original Tag environment has 870 states, while our implementation has 842. The difference arises from the number of terminal states; in the original implementation, they differentiate based on which tile the prey was successfully tagged, while we reduced them to a unique state. The evaluation is conducted by performing 400 runs on 10 initial states (for a total of 4000 runs). The initial states are kept identical for all algorithms.

\section{Additional experiments}
\label{app:additional_exp}
 
Table \ref{tab:appendix_qmdp} shows results for the same RobotDelivery instances as in the main paper, but using the Q-MDP upper bound. Generally, both upper bounds yield similar average results. A notable exception is observed in R-3-1, where AEMS with improved bounds achieves an average return of 1.04, suggesting a shift away from the exit-directly strategy in certain instances. This further underscores the benefit of improving bounds during the online phase.

Table \ref{tab:appendix_qmdp_f.2} presents the results for a modified RobotDelivery version with a movement failure probability  $f=0.2$, also using the Q-MDP upper bound for ease of comparison. AEMS's performance remains unaffected by this change, with variations in return due to the longer expected exit time. AEMS-SR shows lower returns compared to the $f=0.1$ scenario. This outcome was expected given that the increasing $f$ results in expending the support of the beliefs and on increasing the necessary time to pickup and deliver packages. Nonetheless, AEMS-SR still outperforms AEMS and POMCP, both consistently adopting the exit-immediately strategy. 

Table \ref{tab:appendix_qmdp_c.25} details results for the RobotDelivery environment with a state request cost of $c=0.25$, maintaining other parameters as in the main paper. Given our use of a greedy policy to approximate $\policy^*$, increasing the cost requires a more significant reduction in the no-request action's upper bound for AEMS-SR to consider state requests. This makes the problem more difficult. In contrast to POMCP and AEMS, which consistently opt for immediate exits, AEMS-SR still manages to deliver packages in R-3-1 without bounds improvement and in all R-3 instances if updating the offline bounds.

\begin{table*}
\caption{Results for RobotDelivery for POMCP, AEMS and AEMS-SR with 3, 5 and 7 corridors, a time limit of 0.1s, 0.5 and 1s, a cost $c=0.1$, probability of failure of movement $f=0.1$, probability transfer from the waiting area $t=0.8$, expected number of packages $e=3$. POMCP results are averaged on 400 runs. AEMS and AEMS-SR use the Q-MDP upper bound and their results are averaged over 800 runs.  We report the mean and standard error to the mean.}
\label{tab:appendix_qmdp}
\resizebox{\textwidth}{!}{%
\begin{tabular}{cc|c|cc|cc|cc||cc|cc|cc}
\toprule
 &  &  & \multicolumn{6}{c||}{Not Improve Bounds} & \multicolumn{6}{c}{Improve Bounds} \\
 &  & Return & \multicolumn{2}{c|}{Return} & \multicolumn{2}{c|}{NU} & \multicolumn{2}{c||}{ER (\%)} & \multicolumn{2}{c|}{Return} & \multicolumn{2}{c|}{NU} & \multicolumn{2}{c}{ER (\%)} \\
 &  & {\small POMCP} & {\small AEMS} & {\small AEMS-SR} & {\small AEMS} & {\small AEMS-SR} & {\small AEMS} & {\small AEMS-SR} & {\small AEMS} & {\small AEMS-SR} & {\small AEMS} & {\small AEMS-SR} & {\small AEMS} & {\small AEMS-SR} \\
 & T & $\pm0.00$ & $\pm0.00$ & & $\pm0$ & & $\pm0.0$ & & & & $\pm0$ & & &  \\
\midrule
\multirow[c]{3}{*}{R-3} & 0.1 & 0.97 & 0.98 & \textbf{1.37$\pm$0.01} & 56 & \textbf{2473$\pm$3} & 3.9 & \textbf{22.6$\pm$0.4} & 0.98$\pm$0.00 & \textbf{2.33$\pm$0.02} & 56 & \textbf{2209$\pm$7} & 4.0$\pm$0.0 & \textbf{71.6$\pm$0.5} \\
 & 0.5 & 0.97 & 0.98 & \textbf{2.48$\pm$0.03} & 113 & \textbf{10624$\pm$26} & 4.5 & \textbf{42.1$\pm$0.1} & 0.98$\pm$0.00 & \textbf{2.12$\pm$0.02} & 113 & \textbf{7994$\pm$86} & 4.5$\pm$0.0 & \textbf{78.1$\pm$0.5} \\
 & 1.0 & 0.97 & 0.98 & \textbf{2.47$\pm$0.03} & 156 & \textbf{19142$\pm$49} & 4.8 & \textbf{41.7$\pm$0.1} & 1.04$\pm$0.01 & \textbf{2.00$\pm$0.01} & 152 & \textbf{13671$\pm$188} & 15.8$\pm$0.7 & \textbf{79.7$\pm$0.5} \\
\cline{1-15}
\multirow[c]{3}{*}{R-5} & 0.1 & 0.95 & 0.96 & \textbf{1.00$\pm$0.01} & 28 & \textbf{1287$\pm$2} & 3.2 & \textbf{7.7$\pm$0.2} & 0.96$\pm$0.00 & \textbf{2.21$\pm$0.02} & 29 & \textbf{1251$\pm$4} & 3.3$\pm$0.0 & \textbf{53.0$\pm$0.4} \\
 & 0.5 & 0.95 & 0.96 & \textbf{1.01$\pm$0.01} & 58 & \textbf{6637$\pm$8} & 3.8 & \textbf{8.7$\pm$0.2} & 0.96$\pm$0.00 & \textbf{2.26$\pm$0.02} & 59 & \textbf{5918$\pm$28} & 3.8$\pm$0.0 & \textbf{65.9$\pm$0.6} \\
 & 1.0 & 0.95 & 0.96 & \textbf{1.54$\pm$0.01} & 79 & \textbf{13348$\pm$14} & 4.0 & \textbf{35.7$\pm$0.2} & 0.96$\pm$0.00 & \textbf{2.21$\pm$0.02} & 80 & \textbf{10652$\pm$72} & 4.1$\pm$0.0 & \textbf{67.4$\pm$0.6} \\
\cline{1-15}
\multirow[c]{3}{*}{R-7} & 0.1 & 0.93 & \textbf{0.94} & \textbf{0.94$\pm$0.00} & 17 & \textbf{704$\pm$1} & 2.5 & \textbf{6.3$\pm$0.0} & 0.94$\pm$0.00 & \textbf{1.98$\pm$0.02} & 17 & \textbf{713$\pm$2} & 2.6$\pm$0.0 & \textbf{46.3$\pm$0.4} \\
 & 0.5 & 0.93 & \textbf{0.94} & \textbf{0.94$\pm$0.00} & 37 & \textbf{4029$\pm$4} & 3.4 & \textbf{7.0$\pm$0.0} & 0.93$\pm$0.00 & \textbf{2.17$\pm$0.02} & 37 & \textbf{3785$\pm$13} & 3.4$\pm$0.0 & \textbf{53.9$\pm$0.5} \\
 & 1.0 & 0.93 & \textbf{0.94} & \textbf{0.94$\pm$0.00} & 51 & \textbf{8042$\pm$8} & 3.7 & \textbf{7.2$\pm$0.0} & 0.94$\pm$0.00 & \textbf{2.16$\pm$0.02} & 51 & \textbf{7313$\pm$32} & 3.7$\pm$0.0 & \textbf{57.2$\pm$0.5} \\
\cline{1-15}
\bottomrule
\end{tabular}
}%
\end{table*}

\begin{table*}
\caption{Results for RobotDelivery for POMCP, AEMS and AEMS-SR with 3, 5 and 7 corridors, a time limit of 0.1s, 0.5 and 1s, a cost $c=0.1$, probability of failure of movement $\bm{f=0.2}$, probability transfer from the waiting area $t=0.8$, expected number of packages $e=3$. POMCP results are averaged on 400 runs. AEMS and AEMS-SR use the Q-MDP upper bound and their results are averaged over 800 runs.  We report the mean and standard error to the mean.}
\label{tab:appendix_qmdp_f.2}
\resizebox{\textwidth}{!}{%
\begin{tabular}{cc|c|cc|cc|cc||cc|cc|cc}
\toprule
 &  &  & \multicolumn{6}{c||}{Not Improve Bounds} & \multicolumn{6}{c}{Improve Bounds} \\
 &  & Return & \multicolumn{2}{c|}{Return} & \multicolumn{2}{c|}{NU} & \multicolumn{2}{c||}{ER (\%)} & \multicolumn{2}{c|}{Return} & \multicolumn{2}{c|}{NU} & \multicolumn{2}{c}{ER (\%)} \\
 &  & {\small POMCP} & {\small AEMS} & {\small AEMS-SR} & {\small AEMS} & {\small AEMS-SR} & {\small AEMS} & {\small AEMS-SR} & {\small AEMS} & {\small AEMS-SR} & {\small AEMS} & {\small AEMS-SR} & {\small AEMS} & {\small AEMS-SR} \\
 & T & $\pm0.00$ & $\pm0.00$ & & $\pm0$ & &$ \pm0.0$ & & $\pm0.00$ & & $\pm0$ & & $\pm0.0$ & \\
\midrule
\multirow[c]{3}{*}{R-3} & 0.1 & 0.97 & 0.98 & \textbf{1.15$\pm$0.01} & 55 & \textbf{2289$\pm$5} & 3.6 & \textbf{13.6$\pm$0.4} & 0.98 & \textbf{2.43$\pm$0.03} & 55 & \textbf{2240$\pm$7} & 3.6 & \textbf{59.6$\pm$0.5} \\
 & 0.5 & 0.97 & 0.98 & \textbf{2.45$\pm$0.03} & 111 & \textbf{10410$\pm$29} & 4.1 & \textbf{39.5$\pm$0.1} & 0.97 & \textbf{2.33$\pm$0.02} & 112 & \textbf{8627$\pm$66} & 4.1 & \textbf{70.4$\pm$0.5} \\
 & 1.0 & 0.97 & 0.98 & \textbf{2.48$\pm$0.03} & 151 & \textbf{18972$\pm$63} & 4.3 & \textbf{41.8$\pm$0.1} & 0.97 & \textbf{2.23$\pm$0.02} & 152 & \textbf{14465$\pm$149} & 4.3 & \textbf{73.0$\pm$0.5} \\
\cline{1-15}
\multirow[c]{3}{*}{R-5} & 0.1 & 0.94 & 0.95 & \textbf{0.95$\pm$0.00} & 28 & \textbf{1302$\pm$2} & 3.0 & \textbf{6.3$\pm$0.0} & 0.95 & \textbf{0.99$\pm$0.01} & 28 & \textbf{1337$\pm$2} & 3.1 & \textbf{38.5$\pm$0.7} \\
 & 0.5 & 0.95 & 0.95 & \textbf{0.97$\pm$0.00} & 57 & \textbf{6121$\pm$8} & 3.6 & \textbf{7.7$\pm$0.1} & 0.95 & \textbf{2.10$\pm$0.03} & 58 & \textbf{6105$\pm$22} & 3.7 & \textbf{50.4$\pm$0.6} \\
 & 1.0 & 0.95 & 0.95 & \textbf{1.04$\pm$0.01} & 78 & \textbf{11433$\pm$27} & 3.8 & \textbf{11.0$\pm$0.3} & 0.95 & \textbf{2.19$\pm$0.02} & 79 & \textbf{11608$\pm$50} & 3.9 & \textbf{58.9$\pm$0.5} \\
\cline{1-15}
\multirow[c]{3}{*}{R-7} & 0.1 & 0.92 & \textbf{0.93} & \textbf{0.93$\pm$0.00} & 16 & \textbf{725$\pm$1} & 2.4 & \textbf{6.0$\pm$0.0} & \textbf{0.93} & \textbf{0.93$\pm$0.00} & 16 & \textbf{732$\pm$1} & 2.5 & \textbf{6.3$\pm$0.0} \\
 & 0.5 & 0.92 & \textbf{0.93} & \textbf{0.93$\pm$0.00} & 35 & \textbf{4025$\pm$4} & 3.2 & \textbf{7.3$\pm$0.0} & 0.93 & \textbf{1.55$\pm$0.02} & 36 & \textbf{4023$\pm$11} & 3.3 & \textbf{39.3$\pm$0.7} \\
 & 1.0 & 0.92 & \textbf{0.93} & \textbf{0.93$\pm$0.00} & 49 & \textbf{7711$\pm$7} & 3.4 & \textbf{7.7$\pm$0.0} & 0.93 & \textbf{1.89$\pm$0.02} & 50 & \textbf{7553$\pm$27} & 3.5 & \textbf{43.4$\pm$0.6} \\
\cline{1-15}
\bottomrule
\end{tabular}
}%
\end{table*}

\begin{table*}
\caption{Results for RobotDelivery for POMCP, AEMS and AEMS-SR with 3, 5 and 7 corridors, a time limit of 0.1s, 0.5 and 1s, a cost $\bm{c=0.25}$, probability of failure of movement $f=0.2$, probability transfer from the waiting area $t=0.8$, expected number of packages $e=3$. POMCP results are averaged on 400 runs. AEMS and AEMS-SR use the Q-MDP upper bound and their results are averaged over 800 runs.  We report the mean and standard error to the mean.}
\label{tab:appendix_qmdp_c.25}
\resizebox{\textwidth}{!}{%
\begin{tabular}{cc|c|cc|cc|cc||cc|cc|cc}
\toprule
 &  &  & \multicolumn{6}{c||}{Not Improve Bounds} & \multicolumn{6}{c}{Improve Bounds} \\
 &  & Return & \multicolumn{2}{c|}{Return} & \multicolumn{2}{c|}{NU} & \multicolumn{2}{c||}{ER (\%)} & \multicolumn{2}{c|}{Return} & \multicolumn{2}{c|}{NU} & \multicolumn{2}{c}{ER (\%)} \\
 &  & {\small POMCP} & {\small AEMS} & {\small AEMS-SR} & {\small AEMS} & {\small AEMS-SR} & {\small AEMS} & {\small AEMS-SR} & {\small AEMS} & {\small AEMS-SR} & {\small AEMS} & {\small AEMS-SR} & {\small AEMS} & {\small AEMS-SR} \\
 & T  & $\pm0.00$ & $\pm0.00$ & $\pm0.00$ & $\pm0$ & & $\pm0.0$ & & $\pm0.00$ & & $\pm0$ & & $\pm0.0$ & \\
\midrule
\multirow[c]{3}{*}{R-3} & 0.1 & 0.93 & \textbf{0.98} & \textbf{0.98} & 54 & \textbf{2416$\pm$3} & 3.9 & \textbf{6.9$\pm$0.0} & 0.98 & \textbf{2.38$\pm$0.03} & 55 & \textbf{2308$\pm$5} & 3.9 & \textbf{75.7$\pm$0.2} \\
 & 0.5 & 0.94 & \textbf{0.98} & \textbf{0.98} & 110 & \textbf{12043$\pm$14} & 4.5 & \textbf{7.9$\pm$0.0} & 0.98 & \textbf{2.31$\pm$0.03} & 111 & \textbf{11220$\pm$29} & 4.5 & \textbf{78.7$\pm$0.2} \\
 & 1.0 & 0.94 & 0.98 & \textbf{1.65} & 150 & \textbf{24867$\pm$27} & 4.8 & \textbf{37.6$\pm$0.1} & 0.98 & \textbf{2.42$\pm$0.03} & 151 & \textbf{22689$\pm$52} & 4.8 & \textbf{80.8$\pm$0.2} \\
\cline{1-15}
\multirow[c]{3}{*}{R-5} & 0.1 & 0.93 & \textbf{0.96} & \textbf{0.96} & 28 & \textbf{1317$\pm$2} & 3.2 & \textbf{6.7$\pm$0.0} & \textbf{0.96} & \textbf{0.96$\pm$0.00} & 28 & \textbf{1331$\pm$2} & 3.2 & \textbf{6.7$\pm$0.0} \\
 & 0.5 & 0.93 & \textbf{0.96} & \textbf{0.96} & 57 & \textbf{6973$\pm$8} & 3.8 & \textbf{7.8$\pm$0.0} & \textbf{0.96} &\textbf{ 0.96$\pm$0.00} & 59 & \textbf{7051$\pm$7} & 3.8 & \textbf{7.9$\pm$0.0} \\
 & 1.0 & 0.93 & \textbf{0.96} & \textbf{0.96} & 79 & \textbf{14020$\pm$15} & 4.0 & \textbf{8.3$\pm$0.0} & \textbf{0.96} & \textbf{0.96$\pm$0.00} & 80 & \textbf{14185$\pm$13} & 4.1 & \textbf{8.3$\pm$0.0} \\
\cline{1-15}
\multirow[c]{3}{*}{R-7} & 0.1 & 0.92 & \textbf{0.93} & \textbf{0.93} & 17 & \textbf{721$\pm$1} & 2.5 & \textbf{6.3$\pm$0.0} & \textbf{0.94} & 0.93$\pm$0.00 & 17 & \textbf{726$\pm$1} & 2.6 & \textbf{6.4$\pm$0.0} \\
 & 0.5 & 0.92 & \textbf{0.94} & 0.93 & 37 & \textbf{4198$\pm$4} & 3.4 & \textbf{7.7$\pm$0.0} & \textbf{0.94} & \textbf{0.94$\pm$0.00} & 37 & \textbf{4231$\pm$4} & 3.4 & \textbf{7.7$\pm$0.0} \\
 & 1.0 & 0.92 & \textbf{0.94} & \textbf{0.94} & 50 & \textbf{8567$\pm$7} & 3.7 & \textbf{8.2$\pm$0.0} & \textbf{0.94} & \textbf{0.94$\pm$0.00} & 51 & \textbf{8638$\pm$8} & 3.7 & \textbf{8.2$\pm$0.0} \\
\cline{1-15}
\bottomrule
\end{tabular}
}%
\end{table*}

We note that EMS-SR's performance does not uniformly improve with increased compute time per step. Upon examining the results, we identified instances where AEMS-SR, while carrying a package in the corridor, consistently chooses the right action over the down action, resulting in the agent staying in place. This behavior occurs because both actions yield nearly identical optimal values, differing only by a factor of $\gamma$, and that the dynamic programming approach used in backtracking converges to an $\varepsilon$-solution, potentially introducing minor errors that can persist over time due to bound updates.

\begin{remark}
    All AEMS-SR and AEMS experiments were conducted on an Intel Xeon Gold CPU, utilizing a single core and less than 300Mb of RAM.
\end{remark}

\end{document}